%% file: paper.tex
\pdfoutput=1

\documentclass[11pt]{article}

\usepackage{EMNLP2023}

\usepackage{times}
\usepackage{latexsym}
\usepackage{booktabs}
\usepackage{svg}
\usepackage{amsmath}
\usepackage{graphicx}
\usepackage[normalem]{ulem}
\useunder{\uline}{\ul}{}
\usepackage{float}
\usepackage{multirow}
\usepackage{bbold}
\usepackage{hhline}
\usepackage{cleveref}
\usepackage{adjustbox}
\usepackage{todonotes}
\usepackage{listings}
\usepackage{lipsum}
\usepackage[fencedCode]{markdown}
\usepackage[T1]{fontenc}

\usepackage[utf8]{inputenc}

\usepackage{microtype}

\usepackage{xcolor}

\title{Learning From Free-Text Human Feedback \\-- Collect New Datasets Or Extend Existing Ones?}

\author{Dominic Petrak$^{1}$, Nafise Sadat Moosavi$^{2}$, Ye Tian$^{3}$, Nikolai Rozanov$^{3}$, Iryna Gurevych$^{1}$\\
  \\
  $^{1}$UKP Lab, Department of Computer Science, Technical University of Darmstadt, Germany \\
  $^{2}$Department of Computer Science, The University of Sheffield, United Kingdom\\
  $^{3}$Wluper, London, United Kingdom\\  
  \href{https://www.ukp.tu-darmstadt.de}{www.ukp.tu-darmstadt.de}
  }

\begin{document}

\maketitle

\begin{abstract}
Learning from free-text human feedback is essential for dialog systems, but annotated data is scarce and usually covers only a small fraction of error types known in conversational AI. Instead of collecting and annotating new datasets from scratch, recent advances in synthetic dialog generation could be used to augment existing dialog datasets with the necessary annotations. However, to assess the feasibility of such an effort, it is important to know the types and frequency of free-text human feedback included in these datasets. In this work, we investigate this question for a variety of commonly used dialog datasets, including MultiWoZ, SGD, BABI, PersonaChat, Wizards-of-Wikipedia, and the human-bot split of the Self-Feeding Chatbot. Using our observations, we derive new taxonomies for the annotation of free-text human feedback in dialogs and investigate the impact of including such data in response generation for three SOTA language generation models, including GPT-2, LLAMA, and Flan-T5. Our findings provide new insights into the composition of the datasets examined, including error types, user response types, and the relations between them\footnote{Code and data are available on \href{https://github.com/UKPLab/emnlp2023-learning-from-free-text-human-feedback}{GitHub}.}.  

\end{abstract}

\section{Introduction}
\input{introduction.tex}

\section{Related Work}
\input{related_work.tex}

\section{Datasets Examined}\label{sec:datasets}
\input{datasets.tex}

\section{Manual Error Type Analysis and Taxonomies}
\label{sec:collection_error_indicating_sentences}
\input{error_type_analysis.tex}

\section{Automatic Filtering for Potentially Relevant Dialogs}
\label{sec:data_exploration}
\input{data_exploration.tex}

\section{Statistical Analysis}\label{sec:statistical_analysis}
\input{statistical_analysis}

\section{Evaluation and Experiments}
\input{taxonomy_and_experiments}

\section{Discussion}\label{sec:extendibility}
\input{discussion}

\section{Conclusion}
\input{conclusion.tex}

\section{Limitations}
\input{limitations.tex}

\section{Acknowledgments}
This work has been funded by the LOEWE Distinguished Chair \textit{Ubiquitous Knowledge Processing} (LOEWE initiative, Hesse, Germany) and the European Union under the Horizon Europe grant № 101070351 (SERMAS).

\bibliography{anthology,custom}
\bibliographystyle{acl_natbib}

\appendix

\section{Integrated Error Taxonomy -- Details} \label{appendix:integrated_taxonomy}
\input{appendix/details_error_taxonomy.tex}

\section{Error-Indicating Sentences And Phrases}\label{sec:manual_phrase_extraction}
\input{appendix/manual_phrase_extraction.tex}

\section{Automatic Filtering -- Implementation}\label{sec:tfd_impl_details}
\input{appendix/tfd_implementation_details.tex}

\begin{figure*}[htb]
\centering
  \includegraphics[width=0.7\textwidth]{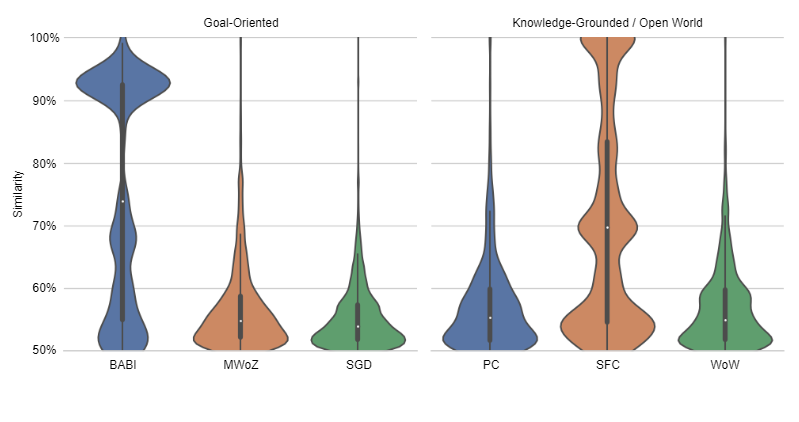}
  \caption{Illustration of the distribution of dialogs with potential real-world error scenarios across datasets.}
  \label{fig:violin_plots}
\end{figure*}

\section{Automatic Filtering -- Sentence-Level Analysis}\label{sec:error_distribution}
\input{appendix/error_distribution.tex}

\section{Task-Oriented Dialogs -- Examples}\label{appendix:dialogue_examples}
\input{appendix/dialogue_examples.tex}

\section{Effectiveness Of Automatic Filtering -- A Detailed Analysis}\label{appendix:effectiveness}
\input{appendix/effectiveness_of_tfd.tex}

\section{Inter-Annotator Agreement -- Detailed Analysis}
\label{appendix:iaa_detailed_analysis}
\input{appendix/class_distribution.tex}

\begingroup
\section{Annotation Guidelines} \label{appendix:annotation_guidelines}
\input{appendix/annotation_guidelines.tex}
\endgroup

\section{Hyperparameters and Baseline Experiments}
\label{appendix:experiments}
\input{appendix/experiments}

\section{Human-Human Dialogs -- Examples}\label{appendix:hh_dialogues}
\input{appendix/polite_answers.tex}

\end{document}

%% file: introduction.tex
If a dialog system generates a dissatisfying or erroneous response, e.g., a response with factually incorrect information, users tend to provide a textual descriptions of what went wrong or what they would have expected~\citep{see-manning-2021-understanding, xu-etal-2023-learning, ung-etal-2022-SaFeRDialogues}. This textual description is usually referred to as \textbf{free-text human feedback}, and it is an important source to improve dialog systems, to keep them engaging and socially acceptable~\citep{shuster2022blenderbot, christiano2023deep, NEURIPS2022_b1efde53, xu-etal-2023-learning}. Learning from such data requires annotations for errors in system utterances and subsequent free-text human feedback, but available datasets are scarce and typically cover only a small subset of the error types known in conversational AI, such as in the cases of  SaFeRDialogues~\citep{ung-etal-2022-SaFeRDialogues} and FITS~\citep{xu-etal-2023-learning}. As a result, data collection is usually a first step for research on learning from free-text human feedback~\citep{hancock-etal-2019-learning, veron2021evaluate, park-etal-2021-scalable}. To avoid this in future research, recent advances in synthetic dialog generation~\citep{kim2022soda, zheng2022augesc} could be used to augment existing dialog datasets with annotations for errors and free-text human feedback. However, to assess the feasibility of such an effort, it is important to know the types and frequency of such data included in these datasets.

In this work, we investigate this question for a variety of dialog datasets, including MultiWoZ~\citep{budzianowski-etal-2018-multiwoz}, PersonaChat~\citep{zhang-etal-2018-personalizing}, Wizards-of-Wikipedia~\citep{dinan2018wizard}, SGD~\citep{rastogi2020towards}, BABI~\citep{bordes2016learning}, and the human-bot split from the Self-Feeding Chatbot~\citep{hancock-etal-2019-learning}. For this, we follow a two-step approach in which we first use the Integrated Error Taxonomy proposed by \citet{higashinaka-etal-2021-integrated} to study the types of errors in system utterances and subsequent user responses in subsets of these datasets. We use the insights gained during this process to derive a new user response type taxonomy and a modified Integrated Error Taxonomy for the annotation of free-text human feedback in dialog data. In the second step, we use Sentence-Transformer~\citep{reimers-gurevych-2019-sentence} to identify similar situations in the remaining dialogs of the datasets. For a subsequent statistical analysis, we manually annotate subsets of the identified dialogs with error and user response types. We also use this manually annotated data to investigate the impact of including this data in response generation, using three SOTA language generation models, i.e., GPT-2~\citep{gpt2}, LLAMA~\citep{touvron2023llama}, and Flan-T5~\citep{flan_t5}.

We find that types and frequency of errors and user responses that include free-text human feedback largely depend on (1) whether the dialogs are human-human or human-bot, and (2) whether the dialogs are task-oriented, open-domain, or knowledge-grounded. Our analysis and experiments show that using our taxonomies is beneficial for identifying free-text human feedback, and that including such data has a positive impact in response generation.

%% file: related_work.tex
\paragraph{Datasets Annotated with Free-Text Human Feedback}\label{sec:existing_datasets}
As of today, dialog datasets with annotations for free-text human feedback are few and mostly focused on specific error types. For example, FITS~\citep{xu-etal-2023-learning}, a dataset of 14,000 human-bot dialogs, provides free-text human feedback for response quality and search query errors. SaFeRDialogues~\citep{ung-etal-2022-SaFeRDialogues} consists of 7,000 human-bot dialogs, but only addresses toxicity in system utterances. 
The dataset published with the Self-Feeding Chatbot~\citep{hancock-etal-2019-learning} provides 60,000 human-bot dialogs, partly annotated with response alternatives for dissatisfying system utterances. Other works, such as \citet{park-etal-2021-scalable} and \citet{veron2021evaluate} collected free-text human feedback for evaluating their approaches, but never made the data publicly available.

In this work, we investigate the types of errors and subsequent user responses in six widely used dialog datasets, such as \mbox{MultiWOZ}~\citep{budzianowski-etal-2018-multiwoz}, PersonaChat~\citep{zhang-etal-2018-personalizing}, and SGD~\citep{rastogi2020towards} to assess their extendibility with annotations for learning from free-text human feedback as an alternative to data collection from scratch.

\paragraph{Error and User Response Type Taxonomies}\label{sec:error_situations}
The error taxonomies used to collect the datasets discussed above are very specific which limits their applicability, e.g., in SaFeRDialogues~\citep{ung-etal-2022-SaFeRDialogues}, they only focus on toxicity. However, errors in conversational AI have long been the subject of research and more comprehensive error taxonomies are already available. For example, \citet{dybkjaer-etal-1996-grice} and \citet{moller2007analysis} focus on errors in task-oriented dialog systems and distinguish multiple error types with focus on content-related errors and practical aspects. More recently published error taxonomies, such as the Integrated Error Taxonomy proposed by \citet{higashinaka-etal-2021-integrated}, distinguish levels of errors, which makes them more broadly applicable. The Integrated Error Taxonomy covers 17 error types in four different levels, including utterance, response, context, and society. Regarding user responses subsequent to errors in system utterances, \citet{see-manning-2021-understanding} proposed a taxonomy that distinguishes dissatisfaction and unclear user utterances.

In this work, we use the Integrated Error Taxonomy by \citet{higashinaka-etal-2021-integrated} to study the errors in the system utterances of 1,200 dialogs from six dialog datasets, based on which we determine (1) the different types of user responses to errors in system utterances, and (2) the limitations of the Integrated Error Taxonomy, such as missing error types. 

%% file: datasets.tex
Table~\ref{tab:datasets} gives an overview of the datasets examined in this work. Overall, we consider six datasets with dialogs of various types, including task-oriented, open-domain, and knowledge-grounded dialogs, as well as human-human and human-bot dialogs.

\begin{table}[ht]
  \centering
  \input{tables/datasets}
\caption{Overview of the datasets examined in this work.}
\label{tab:datasets}
\end{table}

For task-oriented dialog datasets, we consider MultiWoZ~\citep{budzianowski-etal-2018-multiwoz} (\textbf{MWoZ}), SGD~\citep{rastogi2020towards}, and BABI~\citep{bordes2016learning}. They mainly differ in the number of domains included in the dialogs. MWoZ includes seven different domains, SGD 16, and BABI only one (but with dialogs of increasing difficulty). In contrast to MWoZ and SGD, BABI consists of human-bot dialogs. For open-domain dialogs, we consider PersonaChat~\citep{zhang-etal-2018-personalizing} (\textbf{PC}) and the human-bot split of the Self-Feeding Chatbot~\citep{hancock-etal-2019-learning} (\textbf{SFC}). While PC consists of dialogs between two people who are trying to get to know each other, SFC consists of human-bot open-domain dialogs\footnote{SFC is also partially annotated with alternative responses, but we only consider the non-annotated dialogs in this work.}. For knowledge-grounded dialogs, we focus on Wizards-of-Wikipedia~\citep{dinan2018wizard} (\textbf{WoW}), which consists of human-human dialogs.

For simplicity, we do not distinguish between human or bot in the following. We always refer to the utterance of the partner as a system utterance.

%% file: tables/datasets.tex
\resizebox*{\linewidth}{!}{
\begin{tabular}{cccr}
\textbf{Dataset} & \textbf{Type} & \textbf{Mode} & \multicolumn{1}{c}{\textbf{\# Dialogs}} \\ \hline
\multicolumn{1}{l|}{MultiWoZ} & \multicolumn{1}{l|}{Task-Oriented} & \multicolumn{1}{l|}{Human-Human} & 8,483 \\ \hline
\multicolumn{1}{l|}{SGD} & \multicolumn{1}{l|}{Task-Oriented} & \multicolumn{1}{l|}{Human-Human} & 16,000 \\ \hline
\multicolumn{1}{l|}{BABI} & \multicolumn{1}{l|}{Task-Oriented} & \multicolumn{1}{l|}{Human-Bot} & 6,235 \\ \hline
\multicolumn{1}{l|}{PersonaChat} & \multicolumn{1}{l|}{Open-Domain} & \multicolumn{1}{l|}{Human-Human} & 10,907 \\ \hline
\multicolumn{1}{l|}{\begin{tabular}[l]{@{}l@{}}Self-Feeding Chatbot\end{tabular}} & \multicolumn{1}{l|}{Open-Domain} & \multicolumn{1}{l|}{Human-Bot} & 60,000 \\ \hline
\multicolumn{1}{l|}{\begin{tabular}[l]{@{}l@{}}Wizards-of-Wikipedia\end{tabular}} & \multicolumn{1}{l|}{\begin{tabular}[l]{@{}l@{}}Knowledge-\\ Grounded\end{tabular}} & \multicolumn{1}{l|}{Human-Human} & 22,311
\end{tabular}}

%% file: error_type_analysis.tex
We first study the errors in system utterances in a randomly sampled set of $1,200$ dialogs ($200$ dialogs from each of the six datasets), using the Integrated Error Taxonomy proposed by \citet{higashinaka-etal-2021-integrated}. The taxonomy consists of $17$ error types (I1-I17) across four levels: utterance, response, context, and society. They further categorize error types into content violation, i.e., if the error may cause a dialog breakdown, and form violation, i.e., if the system utterance is not interpretable due to massive grammatical problems. Table~\ref{tab:error_type_definition} presents a summary of the error types (see Appendix~\ref{appendix:integrated_taxonomy} for more details).

\begin{table}[ht]
  \centering
  \input{tables/error_type_definition.tex}
\caption{Error Types included in the Integrated Error Taxonomy.}
\label{tab:error_type_definition}
  \end{table}

If we find an error, we analyze the following user response for an \textbf{error-indicating phrase}, a short text fragment of arbitrary length that directly addresses the error in the previous system utterance (e.g., \textit{how do you mean}) or indicates user dissatisfaction (e.g., \textit{I don't like that}), and add the surrounding sentence to our list of unique \textbf{error-indicating sentences}. We use these error-indicating sentences in Section~\ref{sec:data_exploration} to explore the remaining dialogs from each dataset for user responses (and thus errors) that are similar to the ones observed in this step\footnote{We also used the error-indicating phrases instead of the error-indicating-sentences, but found that they are not expressive enough due to their small length (see also Section~\ref{sec:collection_error_indicating_sentences}).}.

Overall, we found $79$ errors in system utterances and collected a set of $67$ error-indicating sentences with an average sentence length of approximately 6.52 words (see Appendix~\ref{sec:manual_phrase_extraction} for all collected phrases and sentences). Each sentence contains a unique error-indicating phrase with an average length of 3.52 words. Contractions (two words that have been connected, e.g., \textit{don't} or \textit{it's}) are considered as one word. Table~\ref{tab:num_of_phrases} shows the distribution of error-indicating sentences across datasets.

\begin{table}[ht]
  \centering
  \input{tables/num_phrases.tex}
\caption{Distribution of error-indicating sentences across datasets. \textit{HH} denotes human-human dialogs and \textit{HB} denotes human-bot dialogs.}
\label{tab:num_of_phrases}
  \end{table}

We find most error-indicating sentences in open-domain and knowledge-grounded datasets, especially in SFC~\citep{hancock-etal-2019-learning}.

\subsection{Modified Integrated Error Taxonomy}\label{sec:error_taxonomy}
\input{manual_error_type_analysis/taxonomy_errors.tex}

\subsection{User Response Type Taxonomy} \label{sec:user_reaction_taxonomy}
\input{manual_error_type_analysis/taxonomy_user_reaction.tex}

%% file: tables/error_type_definition.tex
\resizebox*{\linewidth}{!}{\begin{tabular}{l|l|l}
\multicolumn{1}{c}{\textbf{Level}}   & \multicolumn{1}{c}{\textbf{Form Violation}}                           & \multicolumn{1}{c}{\textbf{Content Violation}}                        \\ \hline
\multirow{2}{*}{Utterance} & Uninterpretable (I1)                                                 & Semantic Error (I3)      \\ \cline{2-3} 
                                    & Grammatical Error (I2)                                               & Wrong Information (I4)   \\ \hline
\multirow{4}{*}{Response}  & Ignore Question (I5)                                                 & Ignore Expectation (I9)  \\ \cline{2-3} 
                                    & Ignore Request (I6)                                                  &                          \\ \cline{2-2}
                                    & Ignore Proposal (I7)                                                 &                          \\ \cline{2-2}
                                    & Ignore Greeting (I8)                                                 &                          \\ \hline
\multirow{3}{*}{Context}   & Unclear Intention (I10)                                              & Self-Contradiction (I13) \\ \cline{2-3} 
                 & \begin{tabular}[c]{@{}l@{}}Topic transition error (I11)\end{tabular} & Contradiction (I14)                                                   \\ \cline{2-3} 
                                    & \begin{tabular}[c]{@{}l@{}}Lack of Information (I12)\end{tabular} & Repetition (I15)         \\ \hline
Society & Lack of Sociality (I16)                                                & \begin{tabular}[c]{@{}l@{}}Lack of Common Sense (I17)\end{tabular}
\end{tabular}}

%% file: tables/num_phrases.tex
\resizebox*{\linewidth}{!}{
\begin{tabular}{c|ccc|cc|c}
\multicolumn{1}{l}{} &
  \multicolumn{3}{c}{\textbf{\begin{tabular}[c]{@{}c@{}}Task-\\ Oriented\end{tabular}}} &
  \multicolumn{2}{c}{\textbf{\begin{tabular}[c]{@{}c@{}}Open-\\ Domain\end{tabular}}} &
  \textbf{\begin{tabular}[c]{@{}c@{}}Know.-\\ Grounded\end{tabular}} \\ \hline
\textbf{Dataset} &
  \begin{tabular}[c]{@{}c@{}}MWoZ\\ (HH)\end{tabular} &
  \begin{tabular}[c]{@{}c@{}}SGD\\ (HH)\end{tabular} &
  \begin{tabular}[c]{@{}c@{}}BABI\\ (HB)\end{tabular} &
  \begin{tabular}[c]{@{}c@{}}PC\\ (HH)\end{tabular} &
  \begin{tabular}[c]{@{}c@{}}SFC\\ (HB)\end{tabular} &
  \begin{tabular}[c]{@{}c@{}}WoW\\ (HH)\end{tabular} \\ \hline
\textbf{\#Sentences} &
  \multicolumn{1}{r}{7} &
  \multicolumn{1}{r}{0} &
  \multicolumn{1}{r|}{5} &
  \multicolumn{1}{r}{9} &
  \multicolumn{1}{r|}{36} &
  \multicolumn{1}{r}{10}
\end{tabular}}

%% file: manual_error_type_analysis/taxonomy_errors.tex
During this study, we found that the Integrated Error Taxonomy~\citep{higashinaka-etal-2021-integrated} has weaknesses. Some error types are never observed and others are missing. Based on these insights, we modify the taxonomy for the classification of errors in system utterances. Table~\ref{tab:improved_taxonomy} shows the result. 

\begin{table}[ht]
  \centering
  \input{tables/improved_taxonomy.tex}
\caption{Modified Integrated Error Taxonomy.}
\label{tab:improved_taxonomy}
  \end{table}

We ignore \textit{Lack of Information} (I12 in Table~\ref{tab:error_type_definition}), since it is rarely observed by \citet{higashinaka-etal-2021-integrated} and we never observed it in our study. For the same reason, we ignore I1-I3. However, we also found them to be rather ambiguous. For example, the \textit{Semantic Error} (I3) is intended to be used for invalid predicate/argument combinations, such as situations where a missing letter results in a different meaning (\textit{raining} instead of \textit{training}). This is similar to the \textit{Lack of Common Sense} error type (I17, now E10), since the model is supposed to be aware of the concept, but not in the given context. For \textit{Wrong Information} (I4), we introduce a new error type, \textit{Factually Incorrect} (E5), that extends the original definition for also taking factually incorrect knowledge into account. Furthermore, we ignore \textit{Contradiction} (I14), since it is covered by \textit{Lack of Common Sense} and \textit{Factually Incorrect} errors. We merge \textit{Ignore Proposal} (I7) and \textit{Ignore Request} (I6) into one error type (E2), since both are very similar in meaning. Next, we merge \textit{Ignore Greeting} (I8) with \textit{Lack of Sociality} (I16, now E9), as the latter implies the first one. We merge \textit{Repetition} (I15) and \textit{Self-Contradiction} (I13) into a new error type, \textit{Conversationality} (E7), since we observed both very rarely and only in situations that the system had lost the thread of the conversation. We also observed instances of incorrectly conveyed attributes (slots) that are not accounted for in the original taxonomy. To address this, we introduce the \textit{Attribute Error} error type (E4).

%% file: tables/improved_taxonomy.tex
\resizebox*{\linewidth}{!}{
\begin{tabular}{l|l|l}
\multicolumn{1}{c}{\textbf{Level}} &
  \multicolumn{1}{c}{\textbf{Error Type}} &
  \multicolumn{1}{c}{\textbf{Description}} \\ \hline
\multirow{5}{*}{Response} &
  Ignore Question (E1) &
  The system utterance ignores the user's question. \\ \cline{2-3} 
 &
  Ignore Request (E2) &
  \begin{tabular}[c]{@{}l@{}}The system utterance ignores the user's request\\ to do something.\end{tabular} \\ \cline{2-3} 
 &
  Ignore Expectation (E3) &
  \begin{tabular}[c]{@{}l@{}}The system utterance does not fulfill the user's\\ expectation.\end{tabular} \\ \cline{2-3} 
 &
  Attribute Error (E4) &
  \begin{tabular}[c]{@{}l@{}}The system utterance suggests that the system\\ did not get the attributes/slots right.\end{tabular} \\ \cline{2-3} 
 &
  Factually Incorrect (E5) &
  \begin{tabular}[c]{@{}l@{}}The system utterance contains information that\\ is factually incorrect.\end{tabular} \\ \hline
\multirow{3}{*}{Context} &
  Topic Transition Error (E6) &
  \begin{tabular}[c]{@{}l@{}}The system utterance transitions to another / a\\ previous topic without reasonable explanation.\end{tabular} \\ \cline{2-3} 
 &
  Conversationality (E7) &
  \begin{tabular}[c]{@{}l@{}}The system utterance indicates that the system lost \\ track, e.g., it repeats previous responses (without \\ asking for missing information) or contradicts itself.\end{tabular} \\ \cline{2-3} 
 &
  Unclear Intention (E8) &
  \begin{tabular}[c]{@{}l@{}}The system utterance suggests that the user's intent\\ was not successfully conveyed.\end{tabular} \\ \hline
\multirow{2}{*}{Society} &
  Lack of Sociality (E9) &
  \begin{tabular}[c]{@{}l@{}}The system utterance lacks consideration of social\\ standards, e.g., greetings, is toxic or  disrespectful.\end{tabular} \\ \cline{2-3} 
 &
  Lack of Common Sense (E10) &
  \begin{tabular}[c]{@{}l@{}}The information in the system utterance opposses\\ the opinion of the majority.\end{tabular}
\end{tabular}}

%% file: manual_error_type_analysis/taxonomy_user_reaction.tex
We observed five different patterns in user responses subsequent to errors in system utterances during this study, which are also reflected in the collected error-indicating sentences. We propose them as a new taxonomy for the annotation of such data in dialogs (Table~\ref{tab:user_response_types}).

\begin{table}[ht]
  \centering
  \input{tables/user_response_types}
\caption{User Response Type Taxonomy.}
\label{tab:user_response_types}
  \end{table}

Among these, we find that UR2, UR3, and UR5 are likely to contain free-text human feedback, such as corrections, new knowledge, or response alternatives. 

%% file: tables/user_response_types.tex
\resizebox*{\linewidth}{!}{\begin{tabular}{ll}
\multicolumn{1}{c}{\textbf{User Response Type}} & \multicolumn{1}{c}{\textbf{Description}} \\ \hline
\multicolumn{1}{l|}{Ignore and Continue (UR1)} & \begin{tabular}[c]{@{}l@{}}The user ignores the error and continues the conversation, \\ e.g., \textit{Okay. Let's leave it like that.}\end{tabular} \\ \hline
\multicolumn{1}{l|}{Repeat or Rephrase (UR2)} & \begin{tabular}[c]{@{}l@{}}The user repeats or rephrases their concern, e.g., \textit{Actually,} \\ \textit{I wanted ...} \end{tabular} \\ \hline
\multicolumn{1}{l|}{\begin{tabular}[c]{@{}l@{}}Make Aware with Correction \\ (UR3)\end{tabular}} & \begin{tabular}[c]{@{}l@{}}The user makes the system aware of the error and provides \\ information to address what is missing or wrong in its \\ utterance, e.g., \textit{No. I wanted you to ...} \end{tabular} \\ \hline
\multicolumn{1}{l|}{\begin{tabular}[c]{@{}l@{}}Make Aware without \\ Correction (UR4)\end{tabular}} & \begin{tabular}[c]{@{}l@{}}The user makes the system aware of the error without \\ providing additional information, e.g., \textit{You're wrong.}\end{tabular} \\ \hline
\multicolumn{1}{l|}{Ask for Clarification (UR5)} & \begin{tabular}[c]{@{}l@{}}The user asks for clarification, e.g., \textit{Are you sure? Is it} \\ \textit{really that ...}\end{tabular}
\end{tabular}}

%% file: data_exploration.tex
Since our study in Section~\ref{sec:collection_error_indicating_sentences} indicated that errors in system utterances are rare, we use Sentence-Transformer~\citep{reimers-gurevych-2019-sentence} to facilitate the process of filtering the remaining dialogs of each datasets for potentially relevant ones, i.e., dialogs with user responses similar to the collected error-indicating sentences.

For each dataset, we decompose every dialog into turns (alternating utterances), extract the user response, and segment it into sentences. Next, we pair these sentences with each of the error-indicating sentences and use a pretrained Sentence-Transformer based on MPNet~\citep{song2020mpnet} to calculate their cosine similarity (see Appendix~\ref{sec:tfd_impl_details} for implementation details). We consider a dialog to be potentially relevant if  at least one of these pairs has a cosine similarity $\geq50\%$. Table~\ref{tab:potential_errors} presents the sizes of the filtered subsets in comparison to the original datasets.

\begin{table}[ht]
  \centering
  \input{tables/potential_errors.tex}
\caption{Size comparison between the filtered subsets and the original datasets. The numbers in brackets show the ratio of relevant dialogs to the original dataset sizes.}
\label{tab:potential_errors}
  \end{table}

With 58.5\%, MWoZ~\citep{budzianowski-etal-2018-multiwoz} contains most of the potentially relevant dialogs. PC~\citep{zhang-etal-2018-personalizing} and WoW~\citep{dinan2018wizard} have the smallest number of such dialogs (8.9\% and 7.57\%, respectively). Overall, only 25\% of the data is potentially relevant, i.e., contains at least one user response that is similar to one of those observed in Section~\ref{sec:collection_error_indicating_sentences}. Hereinafter, we refer to these dialogs as \textbf{filtered dialogs}. We provide a sentence-level analysis in Appendix~\ref{sec:error_distribution}.

%% file: tables/potential_errors.tex
\resizebox*{\linewidth}{!}{\begin{tabular}{c|ccc|cc|c}
\multicolumn{1}{l}{} &
  \multicolumn{3}{c}{\textbf{\begin{tabular}[c]{@{}c@{}}Task-\\ Oriented\end{tabular}}} &
  \multicolumn{2}{c}{\textbf{\begin{tabular}[c]{@{}c@{}}Open-\\ Domain\end{tabular}}} &
  \textbf{\begin{tabular}[c]{@{}c@{}}Know.-\\ Grounded\end{tabular}} \\ \hline
\textbf{Dataset} &
  \begin{tabular}[c]{@{}c@{}}MWoZ\\ (HH)\end{tabular} &
  \begin{tabular}[c]{@{}c@{}}SGD\\ (HH)\end{tabular} &
  \begin{tabular}[c]{@{}c@{}}BABI\\ (HB)\end{tabular} &
  \begin{tabular}[c]{@{}c@{}}PC\\ (HH)\end{tabular} &
  \begin{tabular}[c]{@{}c@{}}SFC\\ (HB)\end{tabular} &
  \begin{tabular}[c]{@{}c@{}}WoW\\ (HH)\end{tabular} \\ \hline
\textbf{\begin{tabular}[c]{@{}c@{}}Original Dialogs\end{tabular}} &
  8,438 &
  16,000 &
  6,235 &
  10,907 &
  60,000 &
  22,311 \\ \hline
\textbf{\begin{tabular}[c]{@{}c@{}}Filtered Dialogs\end{tabular}} &
  \begin{tabular}[c]{@{}c@{}}4,936\\ (58.5\%)\end{tabular} &
  \begin{tabular}[c]{@{}c@{}}5,824\\ (36.4\%)\end{tabular} &
  \begin{tabular}[c]{@{}c@{}}421\\ (6.76\%)\end{tabular} &
  \begin{tabular}[c]{@{}c@{}}974\\ (8.9\%)\end{tabular} &
  \begin{tabular}[c]{@{}c@{}}15,960\\ (26.6\%)\end{tabular} &
  \begin{tabular}[c]{@{}c@{}}1,689\\ (7.57\%)\end{tabular}
\end{tabular}}

%% file: statistical_analysis.tex
In this section, we conduct a statistical analysis of the distribution of error and user response types and their relations in the dialogs of the datasets examined. For this, we manually annotate 555 of the filtered dialogs (100 from each dataset, if available) with error and user response types, using the taxonomies proposed in Section~\ref{sec:error_taxonomy} and \ref{sec:user_reaction_taxonomy}. To avoid bias from our filtering procedure in Section~\ref{sec:data_exploration}, we also consider 600 randomly selected dialogs (100 from each dataset) that were not identified during this process (similarity <50\%) in this analysis. Hereinafter, these dialogs are referred to as \textbf{random dialogs}. In Section~\ref{sub_sec:impact_data_exploration}, we assess the impact of our filtering procedure on this analysis.

Overall, we manually annotate 1,155 dialogs with error and user response types. For annotation, we always consider the entire dialog (the context).

\subsection{Error Type Distribution}\label{sec:error_analysis}
\input{statistical_analysis/error_analysis}

\subsection{User Response Type Distribution}\label{sec:user_reactions}
\input{statistical_analysis/user_reaction_analysis.tex}

\subsection{Relation Between Error and User Response Types}\label{sec:cause_and_effect}
\input{statistical_analysis/cause_and_effect.tex}

\subsection{Impact of Automatic Filtering}\label{sub_sec:impact_data_exploration}
\input{statistical_analysis/impact_automatic_data_exploration}

%% file: statistical_analysis/error_analysis.tex
 Overall, we identified 188 errors across all dialogs. Table~\ref{tab:effectiveness} shows the distribution. 
 
 \begin{table}[ht]
  \centering
  \input{tables/effectiveness_tfd.tex}
\caption{The number of errors in comparison to the number of dialogs considered in this analysis for each dataset.}
\label{tab:effectiveness}
  \end{table}
 
As expected, the filtered dialogs contain a larger number of errors (130 overall) compared to the random dialogs (58 overall), especially for open-domain and knowledge-grounded dialogs, such as SFC~\citep{hancock-etal-2019-learning} and WoW~\citep{dinan2018wizard}. 

\begin{table}[ht]
  \centering
  \input{tables/most_common_errors.tex}
\caption{The most common error types and their frequencies found in both the filtered and random dialogs. The number in brackets shows the ratio to all errors found for the respective dataset.}
\label{tab:most_common_errors}
  \end{table}

Table~\ref{tab:most_common_errors} shows the most common error types and their frequency for both the filtered and random dialogs, which already accounts for 172 of all identified errors\footnote{See \textit{A1} in Table~\ref{tab:class_distribution_tfd_filtered} and \ref{tab:class_distribution_user_responses_tfd_filtered}, Appendix~\ref{appendix:aggregated_errors}, for an aggregated distribution of all errors and user responses.}. In the case of open-domain dialogs, the most frequent error types are \textit{Ignore Question} (E1) and \textit{Topic Transition Error} (E6). This is particularly the case in the SFC dataset~\citep{hancock-etal-2019-learning}, where we find the system utterances to be often out of context. In the case of task-oriented dialogs, \textit{Ignore Request} (E2) and \textit{Ignore Expectation} (E3) are the most common error types. We observe these errors when requests are only partially processed, e.g., when the user requests to book a hotel room and a train, but the system only books the hotel room. Moreover, we find that there is only little variety in language in task-oriented dialogs, regardless of the number of tasks reflected in the dataset (see Appendix~\ref{appendix:dialogue_examples} for examples). In the case of WoW~\citep{dinan2018wizard}, the knowledge-grounded dataset, the \textit{Factually Incorrect} (E5) error is the most frequently observed error type.

%% file: tables/effectiveness_tfd.tex
\resizebox*{\linewidth}{!}{
\begin{tabular}{c|ccc|cc|c}
\multicolumn{1}{l}{} &
  \multicolumn{3}{c}{\textbf{\begin{tabular}[c]{@{}c@{}}Task-\\ Oriented\end{tabular}}} &
  \multicolumn{2}{c}{\textbf{\begin{tabular}[c]{@{}c@{}}Open-\\ Domain\end{tabular}}} &
  \textbf{\begin{tabular}[c]{@{}c@{}}Know.-\\ Grounded\end{tabular}} \\ \hline
\textbf{Dataset} &
  \begin{tabular}[c]{@{}c@{}}MWoZ\\ (HH)\end{tabular} &
  \begin{tabular}[c]{@{}c@{}}SGD\\ (HH)\end{tabular} &
  \begin{tabular}[c]{@{}c@{}}BABI\\ (HB)\end{tabular} &
  \begin{tabular}[c]{@{}c@{}}PC\\ (HH)\end{tabular} &
  \begin{tabular}[c]{@{}c@{}}SFC\\ (HB)\end{tabular} &
  \begin{tabular}[c]{@{}c@{}}WoW\\ (HH)\end{tabular} \\ \hline
\textbf{\begin{tabular}[c]{@{}c@{}}Filtered Dialogs\end{tabular}} & 8/100 & 3/100 & 2/95  & 6/71  & 92/100 & 19/89 \\ \hline
\textbf{\begin{tabular}[c]{@{}c@{}}Random Dialogs\end{tabular}}       & 2/100 & 0/100 & 5/100 & 2/100 & 46/100 & 3/100
\end{tabular}}

%% file: tables/most_common_errors.tex
\resizebox*{\linewidth}{!}{\begin{tabular}{c|rrr|rr|r}
\multicolumn{1}{l}{} &
  \multicolumn{3}{c}{\textbf{\begin{tabular}[c]{@{}c@{}}Task-\\ Oriented\end{tabular}}} &
  \multicolumn{2}{c}{\textbf{\begin{tabular}[c]{@{}c@{}}Open-\\ Domain\end{tabular}}} &
  \multicolumn{1}{c}{\textbf{\begin{tabular}[c]{@{}c@{}}Know.-\\ Grounded\end{tabular}}} \\ \hline
\textbf{Dataset} &
  \multicolumn{1}{c}{\begin{tabular}[c]{@{}c@{}}MWoZ\\ (HH)\end{tabular}} &
  \multicolumn{1}{c}{\begin{tabular}[c]{@{}c@{}}SGD\\ (HH)\end{tabular}} &
  \multicolumn{1}{c|}{\begin{tabular}[c]{@{}c@{}}BABI\\ (HB)\end{tabular}} &
  \multicolumn{1}{c}{\begin{tabular}[c]{@{}c@{}}PC\\ (HH)\end{tabular}} &
  \multicolumn{1}{c|}{\begin{tabular}[c]{@{}c@{}}SFC\\ (HB)\end{tabular}} &
  \multicolumn{1}{c}{\begin{tabular}[c]{@{}c@{}}WoW\\ (HH)\end{tabular}} \\ \hline
\textbf{\begin{tabular}[c]{@{}c@{}}Ignore \\ Question (E1)\end{tabular}}    & 1 (10.0\%)   & -           & 1 (14.3\%) & 1 (12.5\%)  & 67 (48.5\%) & - \\ \hline
\textbf{\begin{tabular}[c]{@{}c@{}}Topic Trans.\\ Error (E6)\end{tabular}}  & -          & -           & -        & 1 (12.5\%)  & 62 (44.9\%) & 4 (18.1\%)\\ \hline
\textbf{\begin{tabular}[c]{@{}c@{}}Factually\\ Incorrect (E5)\end{tabular}} & -          & 2 (66.6\%)    & -        & 1 (12.5\%)  & 3  (2.1\%)& 13 (59.1\%)\\ \hline
\textbf{\begin{tabular}[c]{@{}c@{}}Ignore\\ Expect. (E3)\end{tabular}}      & 2 (20.0\%)   & 1 (33.3\%)    & 1 (14.3\%) & -         & 2  (1.4\%)& 1 (4.5\%)\\ \hline
\textbf{\begin{tabular}[c]{@{}c@{}}Ignore \\ Request (E2)\end{tabular}}     & 3 (30.0\%)   & -           & 1 (14.3\%) & -         & - & -  \\ \hline
\textbf{\begin{tabular}[c]{@{}c@{}}Lack of \\ Sociality (E9)\end{tabular}}  & -          & -           & -        & 2 (25.0\%)  & 3  (2.1\%) & - 
\end{tabular}}

%% file: statistical_analysis/user_reaction_analysis.tex
Table~\ref{tab:user_reactions} shows the distribution of user response types to errors in system utterances for both the random dialogs (R) and the filtered dialogs (F). 

\begin{table}[ht]
  \centering
  \input{tables/user_reactions.tex}
\caption{User response types found in the analyzed dialogs. For the random dialogs, we highlight the user response types that are likely to contain free-text human feedback (Section~\ref{sec:user_reaction_taxonomy}) in bold \textcolor{olive}{green} and the other ones in bold \textcolor{red}{red}.}
\label{tab:user_reactions}
  \end{table}

As described in Section~\ref{sec:user_reaction_taxonomy}, UR2 (\textit{Repeat or Rephrase}), UR3 (\textit{Make Aware with Correction}), and UR5 (\textit{Ask for Clarification}) are likely to contain free-text human feedback. In the case of the filtered dialogs, we find that UR3 and UR5 are more often observed in open-domain and knowledge-grounded dialogs, such as SFC~\citep{hancock-etal-2019-learning} or WoW~\citep{dinan2018wizard}. UR2 is only rarely observed, and only in task-oriented dialogs. However, UR1 (\textit{Ignore and Continue}) is also frequently observed, especially in SFC. For randomly selected dialogs, this is the most frequent user response type (it occurs 42 times).

%% file: tables/user_reactions.tex
\resizebox*{\linewidth}{!}{\begin{tabular}{c|rrrrrr|rrrr|rr}
\multicolumn{1}{l}{} &
  \multicolumn{6}{c}{\textbf{\begin{tabular}[c]{@{}c@{}}Task-\\ Oriented\end{tabular}}} &
  \multicolumn{4}{c}{\textbf{\begin{tabular}[c]{@{}c@{}}Open-\\ Domain\end{tabular}}} &
  \multicolumn{2}{c}{\textbf{\begin{tabular}[c]{@{}c@{}}Know.-\\ Grounded\end{tabular}}} \\ \hline
\multirow{2}{*}{\textbf{Dataset}} &
  \multicolumn{2}{c|}{\begin{tabular}[c]{@{}c@{}}MWoZ\\ (HH)\end{tabular}} &
  \multicolumn{2}{c|}{\begin{tabular}[c]{@{}c@{}}SGD\\ (HH)\end{tabular}} &
  \multicolumn{2}{c|}{\begin{tabular}[c]{@{}c@{}}BABI\\ (HB)\end{tabular}} &
  \multicolumn{2}{c|}{\begin{tabular}[c]{@{}c@{}}PC\\ (HH)\end{tabular}} &
  \multicolumn{2}{c|}{\begin{tabular}[c]{@{}c@{}}SFC\\ (HB)\end{tabular}} &
  \multicolumn{2}{c}{\begin{tabular}[c]{@{}c@{}}WoW\\ (HH)\end{tabular}} \\ \cline{2-13} 
 &
  \multicolumn{1}{c}{F} &
  \multicolumn{1}{c|}{R} &
  \multicolumn{1}{c}{F} &
  \multicolumn{1}{c|}{R} &
  \multicolumn{1}{c}{F} &
  \multicolumn{1}{c|}{R} &
  \multicolumn{1}{c}{F} &
  \multicolumn{1}{c|}{R} &
  \multicolumn{1}{c}{F} &
  \multicolumn{1}{c|}{R} &
  \multicolumn{1}{c}{F} &
  \multicolumn{1}{r}{R} \\ \hline
\textbf{Errors} &
  8 &
  \multicolumn{1}{r|}{2} &
  3 &
  \multicolumn{1}{r|}{0} &
  2 &
  5 &
  6 &
  \multicolumn{1}{r|}{2} &
  92 &
  46 &
  19 &
  3 \\ \hline
\textbf{UR1} &
  1 &
  \multicolumn{1}{r|}{\textcolor{red}{\textbf{2}}} &
  2 &
  \multicolumn{1}{r|}{0} &
  1 &
  \textcolor{red}{\textbf{3}} &
  0 &
  \multicolumn{1}{r|}{\textcolor{red}{\textbf{1}}} &
  4 &
  \textcolor{red}{\textbf{35}} &
  0 &
  \textcolor{red}{\textbf{1}} \\
\textbf{UR2} &
  2 &
  \multicolumn{1}{r|}{0} &
  1 &
  \multicolumn{1}{r|}{0} &
  1 &
  0 &
  0 &
  \multicolumn{1}{r|}{0} &
  0 &
  0 &
  0 &
  0 \\
\textbf{UR3} &
  2 &
  \multicolumn{1}{r|}{0} &
  0 &
  \multicolumn{1}{r|}{0} &
  0 &
  \textcolor{olive}{\textbf{2}} &
  0 &
  \multicolumn{1}{r|}{0} &
  3 &
  \textcolor{olive}{\textbf{1}} &
  9 &
  0 \\
\textbf{UR4} &
  1 &
  \multicolumn{1}{r|}{0} &
  0 &
  \multicolumn{1}{r|}{0} &
  0 &
  0 &
  2 &
  \multicolumn{1}{r|}{\textcolor{red}{\textbf{1}}} &
  34 &
  \textcolor{red}{\textbf{2}} &
  0 &
  \textcolor{red}{\textbf{1}} \\
\textbf{UR5} &
  2 &
  \multicolumn{1}{r|}{0} &
  0 &
  \multicolumn{1}{r|}{0} &
  0 &
  0 &
  4 &
  \multicolumn{1}{r|}{0} &
  51 &
  \textcolor{olive}{\textbf{8}} &
  10 &
  \textcolor{olive}{\textbf{1}}
\end{tabular}}

%% file: statistical_analysis/cause_and_effect.tex
\begin{figure}[htbp]
    \centering
  \includegraphics[width=1.0\linewidth]{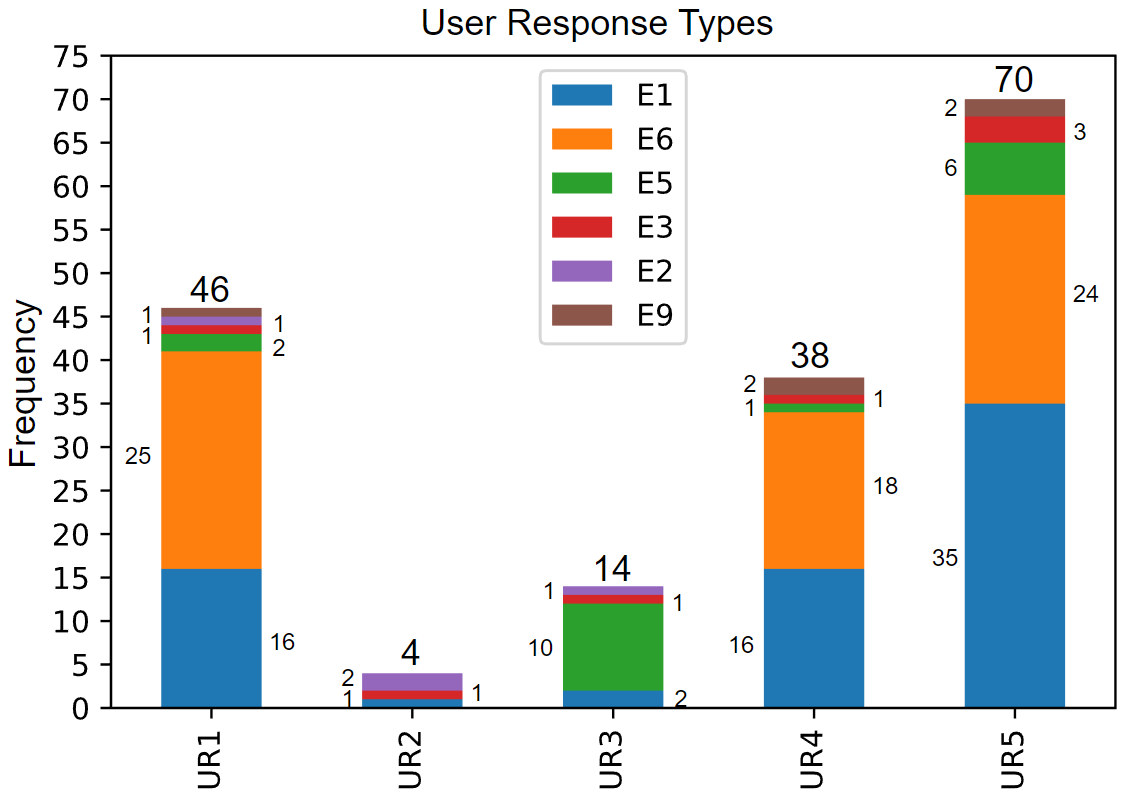}
  \caption{Illustration of the relations between frequent error (E-values) and user response types (UR-values) in both the filtered and random dialogs. The numbers above the bars are the total number of errors for each user response type. The numbers to the left and right of each bar indicate the portion of the respective error type (see color coding).}
  \label{fig:error_reactions}
\end{figure}

Figure~\ref{fig:error_reactions} illustrates the relation between the most common error types (the 172 errors presented in Table~\ref{tab:most_common_errors}) and user response types (Table~\ref{tab:user_reactions}) in both the filtered and random dialogs.

We find that UR1, UR4, and UR5 are the most frequently observed user response types, particularly in the case of \textit{Ignore Question} (E1) and \textit{Topic Transition Error} (E6), which is mostly observed in open-domain datasets (Table~\ref{tab:most_common_errors}). Along with UR3, UR5 is also a frequent response type in the case of \textit{Factually Incorrect} (E5), which is mostly observed in WoW~\citep{dinan2018wizard}. UR2 is only rarely observed. It sometimes occurs in the context of \textit{Ignore Request} (E2) and \textit{Ignore Expectation} (E3), which are mostly found in task-oriented dialogs. 

%% file: statistical_analysis/impact_automatic_data_exploration.tex
As Table~\ref{tab:effectiveness} shows, a total of 188 dialogs was identified to contain errors in this analysis. 130 of them were found in filtered dialogs and 58 in random dialogs (the ones that were missed by our automatic filtering procedure in Section~\ref{sec:data_exploration}). Considering this at the level of user response types (Table~\ref{tab:user_reactions}), 46 of these 58 errors were ignored by users or did not provide any additional information (UR1 or UR4, the ones marked in bold \textcolor{red}{red} in the table), meaning that they are irrelevant because they do not contain free-text human feedback. For the remaining 12 missed errors (UR3 or UR5, the ones marked in bold \textcolor{olive}{green}), we find that they are not reflected in the set of 67 error-indicating sentences used for filtering. Although this limits the effectiveness, we find that our procedure for automatic filtering itself has no negative impact on the results of our analysis, but rather improved annotation efficiency. An approximated recall of $0.72$ supports this assumption. The recall was approximated with respect to the ratio between the size of the filtered subsets and the original datasets\footnote{For calculating the recall, we randomly sampled 25\% of the 555 annotated filtered dialogs, but considered all 600 random dialogs to reflect the proportions from Table~\ref{tab:potential_errors}. We repeated the sampling a thousand times and averaged the recall.} (Table~\ref{tab:potential_errors}) and only considers the 12 missed relevant errors (when considering all missed errors, the recall is $0.35$). We provide a more detailed analysis in Appendix~\ref{appendix:effectiveness}.

%% file: taxonomy_and_experiments.tex
In this section, we use the manually annotated dialogs from Section~\ref{sec:statistical_analysis} in a human evaluation to assess the impact of our modifications to the Integrated Error Taxonomy~\citep{higashinaka-etal-2021-integrated}. We also use these dialogs to investigate the performance impact of errors in system utterances and subsequent user responses as additional input signals for response generation in three SOTA language generation models, including GPT-2~\citep{gpt2}, LLAMA~\citep{touvron2023llama}, and Flan-T5~\citep{flan_t5}.

\subsection{Integrated Error Taxonomy -- Evaluation}\label{sub_sec:taxonomy_evaluation}
\input{taxonomy_and_experiments/credibility}

\subsection{Impact in Response Generation}\label{sub_sec:experiments}
\input{taxonomy_and_experiments/experiments}

%% file: taxonomy_and_experiments/credibility.tex
To evaluate the impact of our modifications to reduce ambiguity and address missing error types in the Integrated Error Taxonomy~\citep{higashinaka-etal-2021-integrated}, we perform a human evaluation. We asked nine experts with NLP background and sound English skills to annotate $600$ dialogs from those that were manually annotated by us in Section~\ref{sec:statistical_analysis} ($300$ from both the filtered and random dialogs, $50$ per dataset) with error and user response types using our modified Integrated Error Taxonomy (Section~\ref{sec:error_taxonomy}) and proposed user response type taxonomy (Section~\ref{sec:user_reaction_taxonomy})\footnote{We provide more details about the annotators (and a more detailed analysis of the results, including edge cases) in Appendix~\ref{appendix:iaa_detailed_analysis}. Please refer to Appendix~\ref{appendix:annotation_guidelines} for the annotation guidelines.}. Each of the dialogs was then assigned to two of these experts and thus annotated three times in total (including our own initial annotation). For comparison, we mapped all annotations back to the original Integrated Error Taxonomy. For merged error types, we asked the annotators for a second assessment using the original taxonomy. Table~\ref{tab:iaa} shows the inter-annotator agreement (IAA) calculated using Krippendorff's Alpha~\citep{krippendorff2004reliability}\footnote{We use the Python library \href{https://github.com/ai-nikolai/annotation_analysis}{annotation\_analysis} for this, last accessed on 15. September 2023.} and summarized by human-human and human-bot dialogs.

\begin{table}[ht]
  \centering
  \input{tables/summarized_iaa.tex}
\caption{Inter-annotator agreement using the Integrated Error Taxonomy \citep{higashinaka-etal-2021-integrated} (\textit{Theirs}) and our modified version (\textit{Ours}).}
\label{tab:iaa}
  \end{table}

In the case of human-human dialogs, the overall agreement is rather low. This also applies to the user response types (what was to be expected, as they directly depend on the error type annotations). During our analysis in Section~\ref{sec:statistical_analysis}, we found that in human-human dialogs, participants tend to suggest disagreement in a friendly manner, which complicates the identification of errors. This is different for human-bot dialogs, where participants tend to provide direct and clear feedback, such as \textit{You're wrong}. We attribute the low agreement for the annotations in human-human dialogs to this observation. Nonetheless, using our modified Integrated Error Taxonomy improves IAA over the original one in all cases. This is most obvious in the case of the human-human filtered dialogs, where it improves IAA by 0.14 points. A detailed analysis revealed that this is mainly due to (1) the condensed number of abstract error types, e.g., we merged ambiguous error types such as \textit{Ignore Proposal} and \textit{Ignore Request}, and (2) the newly added error types, such as \textit{Factually Incorrect}, which were not covered in the original taxonomy.

%% file: tables/summarized_iaa.tex
\resizebox*{0.8\linewidth}{!}{\begin{tabular}{llrrrr}
 &  & \multicolumn{2}{c}{\textbf{Ours}} & \multicolumn{2}{c}{\textbf{Theirs}} \\ \hline
\multicolumn{1}{c}{\textbf{Annotation}} & \multicolumn{1}{c|}{\textbf{Dataset}} & \multicolumn{1}{c}{HH} & \multicolumn{1}{c|}{HB} & \multicolumn{1}{c}{HH} & \multicolumn{1}{c}{HB} \\ \hline
\multicolumn{1}{c|}{\multirow{2}{*}{\textbf{Error Type}}} & \multicolumn{1}{l|}{Filtered} & 0.16 & \multicolumn{1}{r|}{0.91} & 0.02 & 0.89 \\ \cline{2-6} 
\multicolumn{1}{l|}{} & \multicolumn{1}{l|}{Random} & 0.17 & \multicolumn{1}{r|}{0.40} & 0.16 & 0.39 \\ \hline
\multicolumn{1}{c|}{\multirow{2}{*}{\textbf{User Response Type}}} & \multicolumn{1}{l|}{Filtered} & 0.06 & \multicolumn{1}{r|}{0.48} & - & - \\ \cline{2-6} 
\multicolumn{1}{l|}{} & \multicolumn{1}{l|}{Random} & 0.01 & \multicolumn{1}{r|}{0.40} & - & -
\end{tabular}}

%% file: taxonomy_and_experiments/experiments.tex
In the following, we investigate the performance impact of including errors in system utterances and the subsequent user responses as additional input signals in response generation. For this experiment, we consider three different SOTA language generation models: GPT-2~\citep{gpt2}, LLAMA~\citep{touvron2023llama}, and Flan-T5~\citep{flan_t5}. For GPT-2 and Flan-T5, we use the large variants. For LLAMA, we use the 7B variant\footnote{We use the pretrained models available in the \href{https://huggingface.co/models}{Huggingface Model Hub} (last accessed 30. September 2023) for these experiments.}. 

\paragraph{Experimental Setup} We use the dialogs annotated in Section~\ref{sec:statistical_analysis} for this experiment. In a first step, we use the 967 dialogs without error and user response type annotations to train baselines models in the task of response generation. Next, we train the baseline models using the 188 error and user response type annotated dialogs and include the annotated data as additional input signals. For error types, we include the respective system utterance (\textit{Error Text}). For user responses, we include the respective user utterance (\textit{User Response}). As evaluation metrics, we use word-overlapping F1-Score (following \citet{xu-etal-2023-learning} with FITS) and BLEU~\citep{papineni-etal-2002-bleu}. We provide more details, including the baseline results, in Appendix~\ref{appendix:experiments}.

\paragraph{Results} Table~\ref{tab:continuous_experiments} shows the results\footnote{In comparison to the baseline results (Appendix~\ref{appendix:experiments}), we see a performance drop in all configurations. We attribute this to the small number of annotated data and the varying quality.}. We find a large performance gap between Flan-T5~\citep{flan_t5} and the other models. A detailed analysis revealed that both GPT-2~\citep{gpt2} and LLAMA~\citep{touvron2023llama} generate reasonable and fluent responses, but mostly deviate from the target sequence. That aside, including user responses as an additional input signal improves the results over the other configurations, including \textit{None}, for both Flan-T5 and GPT-2. For LLAMA, additionally using the error text improves the results over the other configurations. 

\begin{table}[ht]
  \centering
  \input{tables/error_experiments}
\caption{Experiments with errors in system utterances and subsequent user reactions as additional input signals. For each model, the best performing configuration is highlighted. \textit{Both} includes both feedback signals as additional input signal. \textit{None} was just continually trained on the 188 dialogs, without including the feedback.}
\label{tab:continuous_experiments}
\end{table}

Overall, our results support the findings from recent works on learning from free-text human feedback in that including user responses to errors in system utterances is beneficial~\cite{xu-etal-2023-learning, ung-etal-2022-SaFeRDialogues}. However, we also show that including the error itself can have a positive impact.

%% file: tables/error_experiments.tex
\resizebox*{\linewidth}{!}{\begin{tabular}{crrrrrrrr}
\textbf{} & \multicolumn{2}{c}{\textbf{None}} & \multicolumn{2}{c}{\textbf{Error Text}} & \multicolumn{2}{c}{\textbf{\begin{tabular}[c]{@{}c@{}}User \\ Response\end{tabular}}} & \multicolumn{2}{c}{\textbf{Both}} \\ \hline
\multicolumn{1}{c|}{\textbf{Model}} & \multicolumn{1}{c}{F1} & \multicolumn{1}{c|}{BLEU} & \multicolumn{1}{c}{F1} & \multicolumn{1}{c|}{BLEU} & \multicolumn{1}{c}{F1} & \multicolumn{1}{c|}{BLEU} & \multicolumn{1}{c}{F1} & \multicolumn{1}{c}{BLEU} \\ \hline
\multicolumn{1}{c|}{\textbf{Flan-T5}} & 55.1 & \multicolumn{1}{r|}{35.4} & 54.6 & \multicolumn{1}{r|}{35.3} & \textbf{55.5} & \multicolumn{1}{r|}{\textbf{35.8}} & 55.5 & 35.7 \\ \hline
\multicolumn{1}{c|}{\textbf{GPT-2}} & 19.8 & \multicolumn{1}{r|}{2.9} & 20.8 & \multicolumn{1}{r|}{2.07} & \textbf{20.9} & \multicolumn{1}{r|}{\textbf{2.9}} & 18.7 & 2.8 \\ \hline
\multicolumn{1}{c|}{\textbf{LLAMA}} & 21.9 & \multicolumn{1}{r|}{4.9} & \textbf{24.5} & \multicolumn{1}{r|}{\textbf{10.1}} & 21.9 & \multicolumn{1}{r|}{4.0} & 23.4 & 8.5
\end{tabular}}

%% file: discussion.tex
The goal of this work was to investigate the type and frequency of errors in system utterances and subsequent user responses included in the datasets examined to assess their extendibility with annotations for learning from free-text human feedback. We found that this mostly depends on whether the dialogs are human-human or human-bot. In human-human dialogs, we find that humans rather suggest disagreements in a very polite way instead of accusing the partner of a mistake (see Appendix~\ref{appendix:hh_dialogues} for examples). Accordingly, there is only little free-text human feedback available that could be used for learning (Section~\ref{sec:user_reactions} and \ref{sec:cause_and_effect}). Therefore, it might be hard and ineffective to extend these datasets with annotations for learning from such data. This is different in human-bot dialogs, where humans often react harshly and accusingly to errors in system utterances, resulting in more direct feedback. However, we also found that it depends on the dialog type. In general, we find that open-domain and knowledge-grounded dialogs contain a larger number of errors and user responses that are likely to contain free-text human feedback, making them more suitable for this purpose (Section~\ref{sec:error_analysis}). 

Using the manually annotated dialogs from Section~\ref{sec:statistical_analysis}, our experiments in Section~\ref{sub_sec:experiments} suggest that including user responses to errors in system utterances has a positive impact in response generation, which supports the findings from recent works on including free-text human feedback~\cite{xu-etal-2023-learning, ung-etal-2022-SaFeRDialogues}. Additionally, our results suggest that including the error-annotated system utterance itself can have a positive impact. From our point of view, distinguishing between user response types could be an interesting alternative to binary signals, such as user satisfaction~\citep{hancock-etal-2019-learning} or thumbs-down~\citep{shuster2022blenderbot}, as an indicator of an error in a system utterance. However, the dialogs annotated in Section~\ref{sec:statistical_analysis} do not provide enough such data for a thorough analysis that also takes into account the different types of user responses. Therefore, we leave this as a research question for future work. Our human evaluation in Section~\ref{sub_sec:taxonomy_evaluation} shows that our proposed taxonomies may serve as a promising starting point to obtain the necessary annotaions, although they may not cover all possible error and user response types.

%% file: conclusion.tex
In this work, we examined the dialogs of six datasets from various types, including MultiWoZ, SGD, BABI, PersonaChat, Wizards-of-Wikipedia, and the human-bot split from the Self-Feeding Chatbot, for errors in system utterances and the types of subsequent user responses to assess their extendibility with annotations for learning from free-text human feedback. Our results show that this largely depends on whether the dialogs are human-human or human-bot, and whether they are task-oriented, open-domain, or knowledge-grounded. We found that human-bot dialogs, contain more errors in system utterances that are addressed with free-text human feedback in subsequent user responses, especially in the case of open-domain and knowledge-grounded dialogs. Therefore, it might be feasible to extend these datasets with the needed annotations to support research into methods for learning from free-text human feedback, e.g., by taking advantage of the recent developments in synthetic data generation. We also used the insights gained during this process to propose a new user response type taxonomy and a modified Integrated Error Taxonomy for the annotation of free-text human feedback. Our experiments show that including errors from system utterances and subsequent user responses has a positive impact in response generation.

%% file: limitations.tex
The majority of our evaluation was done manually. Therefore, with respect to the original dataset sizes, we only consider a small fraction of the data in our study. It might be possible that our results would have been clearer when we would have considered more dialogs for the collection of error-indicating sentences. However, our analysis shows that errors found in the randomly selected dialogs are mostly ignored by the user, i.e., the user does not provide free-text human feedback that could be used for learning. Thus, as far as we are concerned, this does not limit the meaningfulness of our results.

Regarding dataset selection, our corpus study (and its results) have only limited expressiveness for knowledge-grounded dialog datasets, since we only consider one of such datasets in our study, Wizards-of-Wikipedia~\citep{dinan2018wizard}. However, this does not affect the relevance of our work, as there are already free-text human feedback annotated datasets available, e.g., FITS~\citep{xu-etal-2023-learning}, and we considered a representative number of datasets from other dialog types for which there is a lack of publicly available feedback-annotated datasets, such as task-oriented dialogs.

The taxonomies used in this work are also subject to limitations. In the case of the modified Integrated Error Taxonomy, our results show that it improves agreement across different dialog types. However, its abstract error types might limit application for specific use cases, e.g., for a more fine-grained consideration of different types of social errors. Moreover, it reflects only error types observed in the datasets examined. The same applies to the user response type taxonomy. 

%% file: appendix/details_error_taxonomy.tex
In this section, we describe the Integrated Error Taxonomy as proposed by \citet{higashinaka-etal-2021-integrated}. In principle, they differentiate between \textit{form violation} and \textit{content violation}. The form violation usually represents errors that oppose some kind of meta criteria, e.g., the form of language or the ignorance of social norms. In contrast, content violations refer to, e.g., inconsistent or redundant utterances, or other things that might cause a dialog breakdown. Content violation is hereinafter abbreviated as \textit{CV} (form violation as \textit{FV}). Furthermore, they generally refer to \textit{utterances}, while we refer to \textit{system utterance}, as this is evident from their examples and simplifies understanding (from our perspective). 

\subsection{Utterance-Level}
Utterance-level errors typically expose language-generation deficiencies of the system.

\begin{itemize}
    \item \textbf{Uninterpretable (I1), FV} -- The system's utterance does not contain recognizable words or is just a text fragment, e.g., the utterance just consists of the words \textit{of yours}.
    \item \textbf{Grammatical Error (I2), FV} -- The system's utterance is grammatical incorrect, e.g., \textit{he will attend yesterday}.
    \item \textbf{Semantic Error (I3), CV} -- The combination of predicates and arguments does not result in any meaning, e.g., \textit{I feel painting.}
    \item \textbf{Wrong Information (I4), CV} -- The given information is clearly wrong, e.g., \textit{Meat Loaf was the songwriter of Jim Steinman} (it was vice versa).
\end{itemize}

\subsection{Response-Level}
Response-level errors describe errors in system utterances, e.g., in responses to user questions or requests.

\begin{itemize}
    \item \textbf{Ignore Question (I5), FV} -- The system utterance ignores the user's question, e.g., \textit{U: What will the weather be like tomorrow? S: Look at these beautiful flowers...}
    \item \textbf{Ignore Request (I6), FV} -- The system ignores the user's request to do something, e.g., \textit{U: Please book me a room in the west part of town. S: I like sports.}
    \item \textbf{Ignore Proposal (I7), FV} -- The system ignores the user's proposal to do something, e.g., \textit{U: Let's talk about movies. S: I like pizza.}
    \item \textbf{Ignore Greeting (I8), FV} -- The system ignores the user's greeting, e.g., \textit{U: Hello, my friend! S: What do you want from me?}
    \item \textbf{Ignore Expectation (I9), CV} -- The system utterance does not have the expected content but the user's intention has been successfully conveyed, e.g., \textit{U: Do you have a favorite football team? S: Yes.} (the system should tell the user the name of his favorite football team).
\end{itemize}

\subsection{Context-Level}
Context-level errors refer not only to the local context, i.e., adjacent pairs of user utterance and system utterance, but to a broader (sometimes global) context.

\begin{itemize}
    \item \textbf{Unclear Intention (I10), FV} -- The system utterance is on the current topic, but the intention is unclear, e.g., \textit{U: What is your favorite movie? S: I like Daddy's Home 2. U: Oh, yes! Mel Gibson is so good in this role... S: The Evil Dead is so bloody...}
    \item \textbf{Topic Transition Error (I11), FV} -- The system suddenly switches to another topic without explanation, e.g., \textit{U: I really enjoyed Feynman's physics lecture. So enlightening! S: Well, what is your favorite food?}
    \item \textbf{Lack of Information (I12), FV} -- The system utterance misses important information, such as subject, object, or modifier, e.g., \textit{U: It's a beautiful day, isn't it? S: I hate.} (object is missing).
    \item \textbf{Self-Contradiction (I13), CV} -- The system utterance contradicts what it said before, e.g., \textit{S: I'm from Boston, Massachusetts. U: Oh, wow! I'm from Bartlesville, Oklahoma! S: Really? I'm from Seattle, Washington!}
    \item \textbf{Contradiction (I14), CV} -- The system utterance contradicts what has been said by the user, e.g., \textit{U: I like strawberries. S: Oh, why don't you like strawberries?}
    \item \textbf{Repetition (I15), CV} -- The system repeats something it has already said before (this includes repeated questioning), e.g., \textit{S: I look forward to the concert. U: Oh, which concert? S: I look forward to the concert.}
\end{itemize}

\subsection{Society-Level}
Society-level errors refer to errors that indicate that the system lacks etiquette or common sense.

\begin{itemize}
    \item \textbf{Lack of Sociality (I16), FV} -- The system utterance lacks consideration towards the user and contains hate speech or is in general disrespectful, e.g., \textit{U: Yesterday I went to the federal employment agency. S: Okay. Are you incapable of finding a job on your own?}
    \item \textbf{Lack of Common Sense (I17), CV} -- The system utterance contains content that differs from the opinion of the majority, e.g., \textit{S: I like to eat chicken raw. U: Oh, dear... You have to cook it first!}
\end{itemize}

%% file: appendix/manual_phrase_extraction.tex
In this section, we present the collected error-indicating sentences along with phrases.

\begin{enumerate}
\item Not really like fandoms, haha Just anything online that people make. (\textbf{Phrase}: \textit{not really like})
\item It is not saturday. (\textbf{Phrase}: \textit{it is not})
\item That doesn't make sense. (\textbf{Phrase}: \textit{doesn't make sense})
\item That makes no sense. (\textbf{Phrase}: \textit{makes no sense})
\item You should put some more things together." (\textbf{Phrase}: \textit{you should})
\item You shouldn't be! (\textbf{Phrase}: \textit{you shouldn't})
\item What do you mean by that?" (\textbf{Phrase}: \textit{what do you mean})
\item What are you talking about? (\textbf{Phrase}: \textit{what are you talking about})
\item It's so important for young people to have diverse interest and develop a wide range of skills, don't you think? (\textbf{Phrase}: \textit{don't you think})
\item I don't know what you're talking about. (\textbf{Phrase}: \textit{don't know})
\item What does that have to do with computer games? (\textbf{Phrase}: \textit{what does that have to do with})
\item Sorry I meant to say for the cat litter. (\textbf{Phrase}: \textit{sorry i meant to say})
\item That didn't have anything to do with school. (\textbf{Phrase}: \textit{didn't have anything to do with})
\item You do not make sense with your response. (\textbf{Phrase}: \textit{your response})
\item That's not what I asked you. (\textbf{Phrase}: \textit{not what i asked})
\item I dont understand. (\textbf{Phrase}: \textit{don't understand})
\item How do you mean? (\textbf{Phrase}: \textit{how do you mean})
\item I don't care about price. (\textbf{Phrase}: \textit{i don't care})
\item You're not answering the questions. (\textbf{Phrase}: \textit{you're not answering})
\item Like I said before I'm not one to read an actual newspaper but I do like reading opinion and political articles. (\textbf{Phrase}: \textit{like i said before})
\item You're not very helpful Help Desk. (\textbf{Phrase}: \textit{not very helpful})
\item Are you sure that there are no hotels on the west side of town? (\textbf{Phrase}: \textit{are you sure})
\item I didn't say anything was scary. (\textbf{Phrase}: \textit{i didn't say})
\item I wouldn't know this. (\textbf{Phrase}: \textit{i wouldn't know this})
\item That sounds too low. (\textbf{Phrase}: \textit{too low})
\item I'm great, but thats off topic. (\textbf{Phrase}: \textit{that's off topic})
\item No, I think when people shape their beards in different ways is really interesting as well! (\textbf{Phrase}: \textit{no, I think})
\item Your doing it wrong my friend. (\textbf{Phrase}: \textit{you're doing it wrong})
\item What are you saying? (\textbf{Phrase}: \textit{what are you saying})
\item At least you have that then. (\textbf{Phrase}: \textit{at least you have})
\item That doesn't answer my question. (\textbf{Phrase}: \textit{that doesn't answer})
\item I am too old to hike I am in my seventies. (\textbf{Phrase}: \textit{i am too old})
\item You aren't staying on topic at all. (\textbf{Phrase}: \textit{not staying on topic})
\item Off the subject, I am thinking of cutting my hair. (\textbf{Phrase}: \textit{off the subject})
\item I'm not ready to book just yet. (\textbf{Phrase}: \textit{i'm not ready})
\item That's not what I asked you. (\textbf{Phrase}: \textit{i asked you})
\item Dude not cool. (\textbf{Phrase}: \textit{dude not cool})
\item I'd really like a 4 star. (\textbf{Phrase}: \textit{i'd really like})
\item Thats nonsense." (\textbf{Phrase}: \textit{thats nonsense})
\item Actually, I apologize no need to book, I was just gathering information." (\textbf{Phrase}: \textit{i apologize})
\item I never said I needed one. (\textbf{Phrase}: \textit{i never said i})
\item No I dont think so. (\textbf{Phrase}: \textit{no i dont think})
\item I didn't mention anything about clowns. (\textbf{Phrase}: \textit{i didn't mention})
\item That is odd for alaska. (\textbf{Phrase}: \textit{that is odd})
\item Not sure what that means? (\textbf{Phrase}: \textit{not sure what that means})
\item It can be what? (\textbf{Phrase}: \textit{it can be what})
\item You should learn! (\textbf{Phrase}: \textit{you should learn})
\item Umm, what? (\textbf{Phrase}: \textit{umm, what})
\item You think so? (\textbf{Phrase}: \textit{you think so})
\item No a park is a place and not a person, (\textbf{Phrase}: \textit{and not})
\item Why do you say that? (\textbf{Phrase}: \textit{why do you say that})
\item I guess I should have asked that first. (\textbf{Phrase}: \textit{i should have asked})
\item I said lets talk about sports. (\textbf{Phrase}: \textit{i said lets talk about})
\item You're being annoying is whats happening. (\textbf{Phrase}: \textit{you're being annoying})
\item You could have stated the goods. (\textbf{Phrase}: \textit{you could have stated})
\item Who was talking about color? (\textbf{Phrase}: \textit{who was talking about})
\item That doesn't really matter. (\textbf{Phrase}: \textit{doesn't really matter})
\item It's actually a 1939 movie that was adapted from a novel written earlier. (\textbf{Phrase}: \textit{it's actually})
\item I don't believe a piano is a stringed instrument. (\textbf{Phrase}: \textit{i don't believe})
\item That's not relevant. (\textbf{Phrase}: \textit{that's not relevant})
\item Check again. (\textbf{Phrase}: \textit{check again})
\item You're wrong. (\textbf{Phrase}: \textit{you're wrong})
\item That doesn't have to do with track. (\textbf{Phrase}: \textit{that doesn't have to do with})
\item Instead could it be in Madrid? (\textbf{Phrase}: \textit{instead could it})
\item I would prefer in Bombay. (\textbf{Phrase}: \textit{i would prefer})
\item No, I don't like that. (\textbf{Phrase}: \textit{i don't like that})
\item No, this does not work for me. (\textbf{Phrase}: \textit{this does not work})
\end{enumerate}

%% file: appendix/tfd_implementation_details.tex
To implement the automatic filtering (Section~\ref{sec:data_exploration}) we use PyTorch~\citep{paszke2019pytorch}, the Transformers library~\citep{wolf-etal-2020-transformers}, and the pretrained \textit{all-mpnet-base-v2} Sentence-Transformer\footnote{\href{https://huggingface.co/sentence-transformers/all-mpnet-base-v2}{Model page} in the HuggingFace Model Hub, last accessed 11. September 2023.}. It is based on MPNet~\citep{song2020mpnet} and finetuned on a large corpus of sentence pairs from multiple tasks and domains, e.g., Yahoo Answers~\citep{zhang2015character} and Reddit Comments~\citep{henderson-etal-2019-repository}, using a contrastive objective. It is a 12-layer Transformer model with a vocabulary size of 30,527 words that calculates the cosine similarity between two sentences in a $768$-dimensional dense vector space. 

Our compute infrastructure consists of one Tesla V100-SXM3 GPU (with 32 GB memory) and it took an average of $76$ mins to run automatic filtering on one dataset.

%% file: appendix/error_distribution.tex
As described in Section~\ref{sec:data_exploration}, we filter on sentence-level for similar user responses. Figure~\ref{fig:violin_plots} illustrates the ranges of similarity between the sentences extracted from the user utterances and the error-indicating sentences, i.e., $50\%-60\%$, $60\%-70\%$, $70\%-80\%$,$80\%-90\%$, $90\%-100\%$. It reflects the share in identified phrases from each of the datasets (see Table~\ref{tab:num_of_phrases}). Most of the phrases were identified in SFC~\citep{hancock-etal-2019-learning}. Only a small amount of phrases came from the other datasets which might be the reason for the clusters in the lower ranges.

%% file: appendix/dialogue_examples.tex
In Section~\ref{sec:error_analysis}, we say that the dialogs of task-oriented datasets considered in our work are of little variety in language and flow. This is most obvious in case of BABI~\citep{bordes2016learning} which we want to show using the following example:

\begin{itemize}
    \item [\textbf{1. USER:}] good morning 
    \item [\textbf{1. SYSTEM:}] hello what can i help you with today
    \item [\textbf{2. USER:}] may i have a table with british cuisine in a cheap price range	
    \item [\textbf{2. SYSTEM:}] i'm on it.  where should it be
    \item [\textbf{3. USER:}] madrid please
    \item [\textbf{3. SYSTEM:}] how many people would be in your party
    \item [\textbf{4. USER:}] for eight please	
    \item [\textbf{4. SYSTEM:}] ok let me look into some options for you    
    \item [\textbf{5. USER:}] actually i would prefer in paris	
    \item [\textbf{5. SYSTEM:}] sure is there anything else to update
    \item [\textbf{6. USER:}] instead could it be for four people	
    \item [\textbf{6. SYSTEM:}] sure is there anything else to update
    \item [\textbf{7. USER:}] instead could it be in a moderate price range	
    \item [\textbf{7. SYSTEM:}] sure is there anything else to update
    \item [\textbf{8. USER:}] no	
    \item [\textbf{8. SYSTEM:}] ok let me look into some options for you
\end{itemize}

This dialog is prototypical for BABI. The first few turns (the first four in this case) contain the original user request. Then, the user asks for changes (turn five to eight). Overall, the dialogs are very scripted. Users are mostly clear and distinctive in their responses. They contain only little variety in language and flow, which makes them seem very unrealistic. We do not observe a change in this regard across tasks. 

The following dialog is from SGD~\citep{rastogi2020towards}:

\begin{itemize}
    \item [\textbf{1. USER:}] Our family is planning for a vacation trip. Please can you find a good rental car for a pleasant trip?
    \item [\textbf{1. SYSTEM:}] At what time do you need the car?
    \item [\textbf{2. USER:}] I can pick up the car at 15:00
    \item [\textbf{2. SYSTEM:}] From when onwards you are looking for the rental car?
    \item [\textbf{3. USER:}] I need the car from 11th of March.
    \item [\textbf{3. SYSTEM:}] So, how long you need the car?
    \item [\textbf{4. USER:}] I might need it till the 13th of this month.
    \item [\textbf{4. SYSTEM:}] Okay! So, where would be your pick up location?
    \item [\textbf{5. USER:}] I can pick up the car in Paris on 9th of this month.
    \item [\textbf{5. SYSTEM:}] Okay! There is a nice car, Chevrolet Cruze, a Standard car is available for pick up on March 9th at Charles de Gaulle International Airport. Do you wish to rent this car?
    \item [\textbf{6. USER:}] Do you have any other cars for I need the car till 10th of this month.
    \item [\textbf{6. SYSTEM:}] Yup! We've 3 more cars. Do you wish to rent Honda Accord, which is a Standard car and its ready for pickup on March 9th from Charles de Gaulle International Airport?
    \item [\textbf{7. USER:}] Okay. I go for the Honda Accord.    
\end{itemize}

The dialog looks much more natural, which might be due to its type. While BABI consists of human-bot dialogs, SGD consists of human-human dialogs. However, the flow is very similar. Up until turn five, the user's request is refined. Then, the system comes up with a suggestion which the user denies (for a couple of turns) by asking for something different. At some point, user's agree or disagree on the system's suggestions. Then the dialog ends. 

The following dialog is from MWoZ~\citep{budzianowski-etal-2018-multiwoz}:

\begin{itemize}
    \item [\textbf{1. USER:}] Hi, I could use help with my planning. I am trying to find a place to stay.
    \item [\textbf{1. SYSTEM:}] Ok, to help us find the right place for you lets talk about what you would like. Do you have a price range you prefer? Or an area of town?
    \item [\textbf{2. USER:}] I would like it to be in the south and have a rating of 3 stars.
    \item [\textbf{2. SYSTEM:}] We have a moderately priced guesthouse and an expensive hotel that match your requirements.
    \item [\textbf{3. USER:}] Does either offer free parking?
    \item [\textbf{3. SYSTEM:}] They both offer free parking and internet.
    \item [\textbf{4. USER:}] Okay can you please book it for 7 people for 4 nights starting on Tuesday?
    \item [\textbf{4. SYSTEM:}] Which one? The guesthouse or the hotel?
    \item [\textbf{5. USER:}] The hotel, please.
    \item [\textbf{5. SYSTEM:}] Reference number is : BS0ODSGJ. Is there anything else?
    \item [\textbf{6. USER:}] Yes, I also need a train from stevenage to cambridge. I want to leave after 10:15 on tuesday.
    \item [\textbf{6. SYSTEM:}] I have train TR0517 that leaves at 11:54. Would that be something that would work for you?
    \item [\textbf{7. USER:}] That's perfect. Could I get the train ID and travel time?
    \item [\textbf{7. SYSTEM:}] The travel time is 49 minutes.
    \item [\textbf{8. USER:}] Great, that's all I need, thanks so much for your help! Have a great day!
    \item [\textbf{8. SYSTEM:}] You're very welcome! Thank you! Bye!
\end{itemize}

Like in case of SGD, the dialog looks very natural. However, in contrast to BABI and SGD, the dialogs consist less of refinements, but more of multiple tasks as shown in this example. First, the user asks for a hotel in a city, then he asks for transportation to that city. Sometimes, he also asks for locations with entertainment. While the first request is usually completed (the booking of a room in this case), the second request is usually about gathering information (the user just asks for the train number, departure and travel time, but not for booking a seat).

%% file: appendix/effectiveness_of_tfd.tex
For the statistical analysis in Section~\ref{sec:statistical_analysis}, we consider $20$ dialogs from each similarity range, i.e., $50\%-60\%$, $60\%-70\%$, $70\%-80\%$, $80\%-90\%$, $90\%-100\%$ (if available, see also Appendix~\ref{sec:error_distribution}) for each dataset examined. As the data in the upper ranges ($80\%-100\%$) is scarce in case of WoW~\citep{dinan2018wizard}, PC~\citep{zhang-etal-2018-personalizing}, and BABI~\citep{bordes2016learning}, the filtered dialogs consists only of $555$ dialogs (instead of $600$ like the randomly selected dialogs). Table~\ref{tab:results_human_evaluation} shows the errors annotated for the statistical analysis with respect to the similarity ranges identified by automatic filtering (meaning that each dialog contains at least one user response with a sentence identified to be similar to at least one error-indicating sentence in this similarity range). \textit{Overall} (O) represents the number of dialogs randomly sampled from the respective similarity range, and \textit{Error} (E) represents the number of dialogs identified in our manual analysis to contain an error in a system utterance.

\begin{table}[ht]
  \centering
  \input{tables/human_evaluation.tex}
\caption{Identified errors in all datasets across similarity ranges.}
\label{tab:results_human_evaluation}
  \end{table}

Overall, only $58$ dialogs of the randomly selected ones ($9.6\%$) contain errors. In the case of automatic filtering, we observe $130$ of such cases. Therefore, automatic filtering shows to facilitate the process of identifying errors in system utterances. Even if the number of identified errors is overall low, most errors are identified in the range of $60\%-100\%$, excluding the densest section in case of MWoZ~\citep{budzianowski-etal-2018-multiwoz}, SGD~\citep{rastogi2020towards}, PC and WoW, $50\%-60\%$ (see also Figure~\ref{fig:violin_plots}).

%% file: tables/human_evaluation.tex
\resizebox*{\linewidth}{!}{\begin{tabular}{ccrrrrrrrrrrrr}
\multicolumn{1}{l}{} &
  \multicolumn{1}{l}{} &
  \multicolumn{6}{c}{\textbf{\begin{tabular}[c]{@{}c@{}}Task-\\ Oriented\end{tabular}}} &
  \multicolumn{4}{c}{\textbf{\begin{tabular}[c]{@{}c@{}}Open-\\ Domain\end{tabular}}} &
  \multicolumn{2}{c}{\textbf{\begin{tabular}[c]{@{}c@{}}Know.-\\ Grounded\end{tabular}}} \\ \hline
\multicolumn{2}{c|}{\textbf{Dataset}} &
  \multicolumn{2}{c}{\begin{tabular}[c]{@{}c@{}}MWoZ\\ (HH)\end{tabular}} &
  \multicolumn{2}{c}{\begin{tabular}[c]{@{}c@{}}SGD\\ (HH)\end{tabular}} &
  \multicolumn{2}{c|}{\begin{tabular}[c]{@{}c@{}}BABI\\ (HB)\end{tabular}} &
  \multicolumn{2}{c}{\begin{tabular}[c]{@{}c@{}}PC\\ (HH)\end{tabular}} &
  \multicolumn{2}{c|}{\begin{tabular}[c]{@{}c@{}}SFC\\ (HB)\end{tabular}} &
  \multicolumn{2}{c}{\begin{tabular}[c]{@{}c@{}}WoW\\ (HH)\end{tabular}} \\ \hline
\multicolumn{2}{l|}{\textbf{Overall / Error}} &
  \multicolumn{1}{c}{O} &
  \multicolumn{1}{c|}{E} &
  \multicolumn{1}{c}{O} &
  \multicolumn{1}{c|}{E} &
  \multicolumn{1}{c}{O} &
  \multicolumn{1}{c|}{E} &
  \multicolumn{1}{c}{O} &
  \multicolumn{1}{c|}{E} &
  \multicolumn{1}{c}{O} &
  \multicolumn{1}{c|}{E} &
  \multicolumn{1}{c}{O} &
  \multicolumn{1}{c}{E} \\ \hline
\multicolumn{1}{c|}{\multirow{6}{*}{\textbf{\begin{tabular}[c]{@{}c@{}}Filtered Dialogs\end{tabular}}}} &
  \multicolumn{1}{c|}{90\% - 100\%} &
  20 &
  \multicolumn{1}{r|}{2} &
  20 &
  \multicolumn{1}{r|}{2} &
  17 &
  \multicolumn{1}{r|}{0} &
  6 &
  \multicolumn{1}{r|}{2} &
  20 &
  \multicolumn{1}{r|}{20} &
  9 &
  4 \\
\multicolumn{1}{c|}{} &
  \multicolumn{1}{c|}{80\% - 90\%} &
  20 &
  \multicolumn{1}{r|}{2} &
  20 &
  \multicolumn{1}{r|}{1} &
  18 &
  \multicolumn{1}{r|}{0} &
  5 &
  \multicolumn{1}{r|}{2} &
  20 &
  \multicolumn{1}{r|}{20} &
  15 &
  9 \\
\multicolumn{1}{c|}{} &
  \multicolumn{1}{c|}{70\% - 80\%} &
  20 &
  \multicolumn{1}{r|}{1} &
  20 &
  \multicolumn{1}{r|}{0} &
  20 &
  \multicolumn{1}{r|}{0} &
  20 &
  \multicolumn{1}{r|}{0} &
  20 &
  \multicolumn{1}{r|}{19} &
  20 &
  4 \\
\multicolumn{1}{c|}{} &
  \multicolumn{1}{c|}{60\% - 70\%} &
  20 &
  \multicolumn{1}{r|}{1} &
  20 &
  \multicolumn{1}{r|}{0} &
  20 &
  \multicolumn{1}{r|}{2} &
  20 &
  \multicolumn{1}{r|}{1} &
  20 &
  \multicolumn{1}{r|}{18} &
  20 &
  2 \\
\multicolumn{1}{c|}{} &
  \multicolumn{1}{c|}{50\% - 60\%} &
  20 &
  \multicolumn{1}{r|}{2} &
  20 &
  \multicolumn{1}{r|}{0} &
  20 &
  \multicolumn{1}{r|}{0} &
  20 &
  \multicolumn{1}{r|}{1} &
  20 &
  \multicolumn{1}{r|}{15} &
  20 &
  0 \\ \cline{2-14} 
\multicolumn{1}{c|}{} &
  \multicolumn{1}{c|}{Overall} &
  100 &
  \multicolumn{1}{r|}{8} &
  100 &
  \multicolumn{1}{r|}{3} &
  95 &
  \multicolumn{1}{r|}{2} &
  71 &
  \multicolumn{1}{r|}{6} &
  100 &
  \multicolumn{1}{r|}{92} &
  89 &
  19 \\ \hline
\multicolumn{1}{c|}{\textbf{\begin{tabular}[c]{@{}c@{}}Random Dialogs\end{tabular}}} &
  \multicolumn{1}{l|}{} &
  100 &
  \multicolumn{1}{r|}{2} &
  100 &
  \multicolumn{1}{r|}{0} &
  100 &
  \multicolumn{1}{r|}{5} &
  100 &
  \multicolumn{1}{r|}{2} &
  100 &
  \multicolumn{1}{r|}{46} &
  100 &
  3
\end{tabular}}

%% file: appendix/class_distribution.tex
This section gives more insights on the inter-annotator agreement. All additional annotators that participated in this study were non-native speakers. They were experts from our lab with sound English skills and NLP background. We did not select them based on specific criteria; they participated voluntarily. Accordingly, they were not paid extra for this, since they did the annotations during their working hours. For annotation, we did not use any specific tool. We provided the annotators with dialogs in json format and asked them to do the annotations directly in the respective files. See Section~\ref{appendix:annotation_guidelines} for the annotation guidelines. Table~\ref{tab:_iaa} shows the inter-annotator agreement for each dataset using our modified Integrated Error Taxonomy.

\begin{table}[ht]
  \centering
  \input{tables/inter_annotator.tex}
\caption{Inter-annotator agreement for each dataset.}
\label{tab:_iaa}
  \end{table}

In the case of human-human dialogs, the overall agreement is rather low (except for PersonaChat~\citep{zhang-etal-2018-personalizing}). We find that errors are hard to identify in these dialogs, as humans rather suggest disagreements instead of accusing the partner of a mistake. This is also reflected in the user response type agreement since it depends on the error type annotation. However, PersonaChat seems to be different (according to Table~\ref{tab:most_common_errors}). We attribute this to the dialog type, which is open-domain, where we find that humans react harshly and accusing to errors in system utterances, resulting in more direct feedback that is easier to identify. 

\begin{table}[ht]
  \centering
  \input{tables/improved_inter_annotator.tex}
\caption{Inter-annotator-agreement for the \citet{higashinaka-etal-2021-integrated} taxonomy.}
\label{tab:_improved_iaa}
  \end{table}
  
Table~\ref{tab:_improved_iaa} shows the inter-annotator agreement for each dataset using the original Integrated Error Taxonomy as proposed by \citet{higashinaka-etal-2021-integrated}. Using this taxonomy deteriorates the inter-annotator agreement. This is most obvious in case of MWoZ~\citep{budzianowski-etal-2018-multiwoz} and  PC~\citep{zhang-etal-2018-personalizing}, which are both human-human datasets. A detailed analysis revealed that this is mostly due to over-specialized error types which were merged in our modified taxonomy, such as \textit{ignore expectation} and \textit{ignore request}, I9 and I6 in the original taxonomy (Table~\ref{tab:error_type_definition}). Another reason are the newly added error types, such as \textit{factually incorrect}, E5 (Table~\ref{tab:improved_taxonomy}), which were not covered in the original taxonomy, but occur in the dialogs.

\subsection{Edge Cases}\label{appendix:aggregated_errors}
Table~\ref{tab:class_distribution_tfd_filtered} shows the aggregated error type distribution (error type annotation from both the filtered and random subsets). \textit{A1} denotes the authors' annotations done for the statistical analysis in Section~\ref{sec:statistical_analysis}.

\begin{table}[ht]
  \centering
  \input{tables/class_distribution_tfd_filtered.tex}
\caption{Error types in both the filtered and random dialogs.}
\label{tab:class_distribution_tfd_filtered}
  \end{table}

Overall, the distribution is very broadly spread. However, in most cases, it seems like at least two annotators agree. There are only a few outliers where there is a large deviation, i.e., unclear intention (E8 in Table~\ref{tab:improved_taxonomy}) in case of MWoZ~\citep{budzianowski-etal-2018-multiwoz} and WoW~\citep{dinan2018wizard}, topic transition error (E6) in case of PC~\citep{zhang-etal-2018-personalizing}, factually incorrect (E5) and attribute error (E4) in case of WoW. For example, attribute error is defined as an error type that rather addresses task-oriented dialogs, but annotator two found it five times in the WoW dataset. During our analysis, we found that factually incorrect would have described these cases more accurately. In the case of unclear intention in WoW and MWoZ, we found that annotator two marked some cases as errors that are actually not necessarily errors. The same applies to the factually incorrect errors in BABI~\cite{bordes2016learning} (which consists of task-oriented dialogs). In the case of PC, we found that topic transition error is in most cases the most obvious error type, and in our opinion, annotator three was right in most of the cases. 

In summary, we find that deviations are primarily the result of (1) how the annotators interpret the descriptions of the error types (based on their experience), and (2) biases in the data. The former could probably be addressed by more examples in the annotation guidelines. The latter is a bit more difficult. In these cases, a multi-step annotation process could be useful, where annotators mark errors they are not sure about to be discussed before they are finally annotated.

\begin{table}[ht]
  \centering
  \input{tables/class_distribution_user_responses_tfd_filtered.tex}
\caption{User response types in both the filtered and random dialogs.}
\label{tab:class_distribution_user_responses_tfd_filtered}
  \end{table}

Table~\ref{tab:class_distribution_user_responses_tfd_filtered} shows the aggregated distribution of user response types. What stands out here are the differences in the situations where users ignore the error situations (UR1 in Table~\ref{tab:user_response_types}) in the case of PC (annotator three) and WoW (annotator two). However, we found no pattern in the underlying error situations. We suspect that this is due to the nature of human-human dialogs (PC and WoW consist of human-human dialogs). As we have pointed out before, humans behave more cautiously and politely in human-human dialogs, while in contrast they clearly point out errors in human-bot dialogs (see Appendix~\ref{appendix:hh_dialogues} for examples).

%% file: tables/inter_annotator.tex
\resizebox*{\linewidth}{!}{\begin{tabular}{clrrrrrr}
\multicolumn{1}{l}{} &
   &
  \multicolumn{3}{c}{\textbf{\begin{tabular}[c]{@{}c@{}}Task-\\ Oriented\end{tabular}}} &
  \multicolumn{2}{c}{\textbf{\begin{tabular}[c]{@{}c@{}}Open-\\ Domain\end{tabular}}} &
  \multicolumn{1}{c}{\textbf{\begin{tabular}[c]{@{}c@{}}Know.-\\ Grounded\end{tabular}}} \\ \hline
\multicolumn{2}{c|}{\textbf{Dataset}} &
  \multicolumn{1}{c}{\begin{tabular}[c]{@{}c@{}}MWoZ\\ (HH)\end{tabular}} &
  \multicolumn{1}{c}{\begin{tabular}[c]{@{}c@{}}SGD\\ (HH)\end{tabular}} &
  \multicolumn{1}{c|}{\begin{tabular}[c]{@{}c@{}}BABI\\ (HB)\end{tabular}} &
  \multicolumn{1}{c}{\begin{tabular}[c]{@{}c@{}}PC\\ (HH)\end{tabular}} &
  \multicolumn{1}{c|}{\begin{tabular}[c]{@{}c@{}}SFC\\ (HB)\end{tabular}} &
  \multicolumn{1}{c}{\begin{tabular}[c]{@{}c@{}}WoW\\ (HH)\end{tabular}} \\ \hline
\multicolumn{1}{c|}{\multirow{2}{*}{\textbf{\begin{tabular}[c]{@{}c@{}}Error\\ Type\end{tabular}}}} &
  \multicolumn{1}{l|}{Filtered} &
  0.01 &
  0.0 &
  \multicolumn{1}{r|}{1.0} &
  0.51 &
  \multicolumn{1}{r|}{0.81} &
  0.12 \\ \cline{2-8} 
\multicolumn{1}{c|}{} &
  \multicolumn{1}{l|}{Random} &
  0.55 &
  0.01 &
  \multicolumn{1}{r|}{-0.01} &
  0.09 &
  \multicolumn{1}{r|}{0.80} &
  0.02 \\ \hline
\multicolumn{1}{c|}{\multirow{2}{*}{\textbf{\begin{tabular}[c]{@{}c@{}}User\\ Res. Type\end{tabular}}}} &
  \multicolumn{1}{l|}{Filtered} &
  0.04 &
  0.0 &
  \multicolumn{1}{r|}{0.23} &
  0.16 &
  \multicolumn{1}{r|}{0.72} &
  0.04 \\ \cline{2-8} 
\multicolumn{1}{c|}{} &
  \multicolumn{1}{l|}{Random} &
  0.05 &
  0.0 &
  \multicolumn{1}{r|}{0.0} &
  0.01 &
  \multicolumn{1}{r|}{0.79} &
  -0.02
\end{tabular}
}

%% file: tables/improved_inter_annotator.tex
\resizebox*{\linewidth}{!}{\begin{tabular}{clcccccc}
\multicolumn{1}{l}{} &
   &
  \multicolumn{3}{c}{\textbf{\begin{tabular}[c]{@{}c@{}}Task-\\ Oriented\end{tabular}}} &
  \multicolumn{2}{c}{\textbf{\begin{tabular}[c]{@{}c@{}}Open-\\ Domain\end{tabular}}} &
  \textbf{\begin{tabular}[c]{@{}c@{}}Know.-\\ Grounded\end{tabular}} \\ \hline
\multicolumn{2}{c|}{\textbf{Dataset}} &
  \begin{tabular}[c]{@{}c@{}}MWoZ\\ (HH)\end{tabular} &
  \begin{tabular}[c]{@{}c@{}}SGD\\ (HH)\end{tabular} &
  \multicolumn{1}{c|}{\begin{tabular}[c]{@{}c@{}}BABI\\ (HB)\end{tabular}} &
  \begin{tabular}[c]{@{}c@{}}PC\\ (HH)\end{tabular} &
  \multicolumn{1}{c|}{\begin{tabular}[c]{@{}c@{}}SFC\\ (HB)\end{tabular}} &
  \begin{tabular}[c]{@{}c@{}}WoW\\ (HH)\end{tabular} \\ \hline
\multicolumn{1}{c|}{\multirow{2}{*}{\textbf{\begin{tabular}[c]{@{}c@{}}Error\\ Type\end{tabular}}}} &
  \multicolumn{1}{l|}{Filtered} &
  \begin{tabular}[c]{@{}c@{}}-0.10\\ (-0.11)\end{tabular} &
  \begin{tabular}[c]{@{}c@{}}0.0\\ (-0.0)\end{tabular} &
  \multicolumn{1}{c|}{\begin{tabular}[c]{@{}c@{}}1.0\\ (-0.0)\end{tabular}} &
  \begin{tabular}[c]{@{}c@{}}0.26\\ (-0.25)\end{tabular} &
  \multicolumn{1}{c|}{\begin{tabular}[c]{@{}c@{}}0.80\\ (-0.01)\end{tabular}} &
  \begin{tabular}[c]{@{}c@{}}-0.09\\ (-0.21)\end{tabular} \\ \cline{2-8} 
\multicolumn{1}{c|}{} &
  \multicolumn{1}{l|}{Random} &
  \begin{tabular}[c]{@{}c@{}}0.55\\ (-0.0)\end{tabular} &
  \begin{tabular}[c]{@{}c@{}}0.01\\ (-0.0)\end{tabular} &
  \multicolumn{1}{c|}{\begin{tabular}[c]{@{}c@{}}-0.01\\ (-0.0)\end{tabular}} &
  \begin{tabular}[c]{@{}c@{}}0.09\\ (-0.0)\end{tabular} &
  \multicolumn{1}{c|}{\begin{tabular}[c]{@{}c@{}}0.80\\ (-0.01)\end{tabular}} &
  \begin{tabular}[c]{@{}c@{}}0.0\\ (-0.02)\end{tabular}
\end{tabular}}

%% file: tables/class_distribution_tfd_filtered.tex
\resizebox*{\linewidth}{!}{
\begin{tabular}{c|rrrrrrrrr|rrrrrr|rrr}
\multicolumn{1}{l}{} &
  \multicolumn{9}{c}{\textbf{\begin{tabular}[c]{@{}c@{}}Task-\\ Oriented\end{tabular}}} &
  \multicolumn{6}{c}{\textbf{\begin{tabular}[c]{@{}c@{}}Open-\\ Domain\end{tabular}}} &
  \multicolumn{3}{c}{\textbf{\begin{tabular}[c]{@{}c@{}}Knowledge-\\ Grounded\end{tabular}}} \\ \hline
\multicolumn{1}{l|}{} &
  \multicolumn{3}{c|}{\begin{tabular}[c]{@{}c@{}}MWoZ\\ (HH)\end{tabular}} &
  \multicolumn{3}{c|}{\begin{tabular}[c]{@{}c@{}}SGD\\ (HH)\end{tabular}} &
  \multicolumn{3}{c|}{\begin{tabular}[c]{@{}c@{}}BABI\\ (HB)\end{tabular}} &
  \multicolumn{3}{c|}{\begin{tabular}[c]{@{}c@{}}PC\\ (HH)\end{tabular}} &
  \multicolumn{3}{c|}{\begin{tabular}[c]{@{}c@{}}SFC\\ (HB)\end{tabular}} &
  \multicolumn{3}{c}{\begin{tabular}[c]{@{}c@{}}WoW\\ (HH)\end{tabular}} \\ \hline
\textbf{Annotator} &
  \multicolumn{1}{l}{A1} &
  \multicolumn{1}{l}{A2} &
  \multicolumn{1}{l|}{A3} &
  \multicolumn{1}{l}{A1} &
  \multicolumn{1}{l}{A2} &
  \multicolumn{1}{l|}{A3} &
  \multicolumn{1}{l}{A1} &
  \multicolumn{1}{l}{A2} &
  \multicolumn{1}{l|}{A3} &
  \multicolumn{1}{l}{A1} &
  \multicolumn{1}{l}{A2} &
  \multicolumn{1}{l|}{A3} &
  \multicolumn{1}{l}{A1} &
  \multicolumn{1}{l}{A2} &
  \multicolumn{1}{l|}{A3} &
  \multicolumn{1}{l}{A1} &
  \multicolumn{1}{l}{A2} &
  \multicolumn{1}{l}{A3} \\ \hline
\textbf{\begin{tabular}[c]{@{}c@{}}Ignore\\ Question (E1)\end{tabular}} &
  1 &
  2 &
  \multicolumn{1}{r|}{2} &
  - &
  - &
  \multicolumn{1}{r|}{1} &
  1 &
  2 &
  1 &
  1 &
  2 &
  \multicolumn{1}{r|}{5} &
  67 &
  64 &
  66 &
  - &
  1 &
  3 \\ \hline
\textbf{\begin{tabular}[c]{@{}c@{}}Ignore\\ Request (E2)\end{tabular}} &
  3 &
  3 &
  \multicolumn{1}{r|}{3} &
  - &
  - &
  \multicolumn{1}{r|}{-} &
  1 &
  - &
  - &
  - &
  - &
  \multicolumn{1}{r|}{-} &
  1 &
  9 &
  6 &
  - &
  - &
  - \\ \hline
\textbf{\begin{tabular}[c]{@{}c@{}}Ignore\\ Expect. (E3)\end{tabular}} &
  2 &
  3 &
  \multicolumn{1}{r|}{3} &
  1 &
  - &
  \multicolumn{1}{r|}{1} &
  1 &
  1 &
  1 &
  - &
  - &
  \multicolumn{1}{r|}{-} &
  2 &
  1 &
  - &
  2 &
  - &
  - \\ \hline
\textbf{\begin{tabular}[c]{@{}c@{}}Attribute \\ Error (E4)\end{tabular}} &
  3 &
  1 &
  \multicolumn{1}{r|}{-} &
  - &
  - &
  \multicolumn{1}{r|}{-} &
  4 &
  - &
  3 &
  - &
  2 &
  \multicolumn{1}{r|}{3} &
  - &
  3 &
  - &
  1 &
  5 &
  1 \\ \hline
\textbf{\begin{tabular}[c]{@{}c@{}}Factually\\ Incorrect (E5)\end{tabular}} &
  - &
  2 &
  \multicolumn{1}{r|}{-} &
  2 &
  - &
  \multicolumn{1}{r|}{-} &
  - &
  4 &
  - &
  1 &
  2 &
  \multicolumn{1}{r|}{-} &
  3 &
  1 &
  - &
  13 &
  1 &
  1 \\ \hline
\textbf{\begin{tabular}[c]{@{}c@{}}Topic Trans.\\ Error (E6)\end{tabular}} &
  - &
  - &
  \multicolumn{1}{r|}{1} &
  - &
  - &
  \multicolumn{1}{r|}{-} &
  - &
  - &
  - &
  2 &
  2 &
  \multicolumn{1}{r|}{10} &
  62 &
  58 &
  58 &
  4 &
  - &
  1 \\ \hline
\textbf{Convers. (E7)} &
  1 &
  - &
  \multicolumn{1}{r|}{2} &
  - &
  - &
  \multicolumn{1}{r|}{1} &
  - &
  - &
  - &
  1 &
  1 &
  \multicolumn{1}{r|}{1} &
  - &
  - &
  2 &
  1 &
  - &
  3 \\ \hline
\textbf{\begin{tabular}[c]{@{}c@{}}Unclear\\ Intention (E8)\end{tabular}} &
  - &
  12 &
  \multicolumn{1}{r|}{-} &
  - &
  - &
  \multicolumn{1}{r|}{-} &
  - &
  - &
  - &
  - &
  - &
  \multicolumn{1}{r|}{1} &
  - &
  2 &
  2 &
  - &
  13 &
  - \\ \hline
\textbf{\begin{tabular}[c]{@{}c@{}}Lack of\\ Sociality (E9)\end{tabular}} &
  - &
  - &
  \multicolumn{1}{r|}{-} &
  - &
  - &
  \multicolumn{1}{r|}{-} &
  - &
  - &
  - &
  2 &
  1 &
  \multicolumn{1}{r|}{4} &
  3 &
  2 &
  1 &
  - &
  - &
  - \\ \hline
\textbf{\begin{tabular}[c]{@{}c@{}}Lack of \\ Com. Sense\\ (E10)\end{tabular}} &
  - &
  - &
  \multicolumn{1}{r|}{1} &
  - &
  - &
  \multicolumn{1}{r|}{-} &
  - &
  - &
  - &
  1 &
  2 &
  \multicolumn{1}{r|}{2} &
  - &
  - &
  - &
  1 &
  - &
  1
\end{tabular}}

%% file: tables/class_distribution_user_responses_tfd_filtered.tex
\resizebox*{\linewidth}{!}{
\begin{tabular}{c|rrrrrrrrr|rrrrrr|rrr}
\multicolumn{1}{l}{} &
  \multicolumn{9}{c}{\textbf{\begin{tabular}[c]{@{}c@{}}Task-\\ Oriented\end{tabular}}} &
  \multicolumn{6}{c}{\textbf{\begin{tabular}[c]{@{}c@{}}Open-\\ Domain\end{tabular}}} &
  \multicolumn{3}{c}{\textbf{\begin{tabular}[c]{@{}c@{}}Knowledge-\\ Grounded\end{tabular}}} \\ \hline
\multicolumn{1}{l|}{} &
  \multicolumn{3}{c|}{\begin{tabular}[c]{@{}c@{}}MWoZ\\ (HH)\end{tabular}} &
  \multicolumn{3}{c|}{\begin{tabular}[c]{@{}c@{}}SGD\\ (HH)\end{tabular}} &
  \multicolumn{3}{c|}{\begin{tabular}[c]{@{}c@{}}BABI\\ (HB)\end{tabular}} &
  \multicolumn{3}{c|}{\begin{tabular}[c]{@{}c@{}}PC\\ (HH)\end{tabular}} &
  \multicolumn{3}{c|}{\begin{tabular}[c]{@{}c@{}}SFC\\ (HB)\end{tabular}} &
  \multicolumn{3}{c}{\begin{tabular}[c]{@{}c@{}}WoW\\ (HH)\end{tabular}} \\ \hline
\textbf{Annotator} &
  \multicolumn{1}{l}{A1} &
  \multicolumn{1}{l}{A2} &
  \multicolumn{1}{l|}{A3} &
  \multicolumn{1}{l}{A1} &
  \multicolumn{1}{l}{A2} &
  \multicolumn{1}{l|}{A3} &
  \multicolumn{1}{l}{A1} &
  \multicolumn{1}{l}{A2} &
  \multicolumn{1}{l|}{A3} &
  \multicolumn{1}{l}{A1} &
  \multicolumn{1}{l}{A2} &
  \multicolumn{1}{l|}{A3} &
  \multicolumn{1}{l}{A1} &
  \multicolumn{1}{l}{A2} &
  \multicolumn{1}{l|}{A3} &
  \multicolumn{1}{l}{A1} &
  \multicolumn{1}{l}{A2} &
  \multicolumn{1}{l}{A3} \\ \hline
\textbf{UR1} & 3 & 8 & \multicolumn{1}{r|}{5} & 2 & - & \multicolumn{1}{r|}{3} & 4 & 2 & 3 & 1 & 2 & \multicolumn{1}{r|}{23} & 39 & 40 & 36 & 1 & 6 & 2 \\ \hline
\textbf{UR2} & 2 & 9 & \multicolumn{1}{r|}{3} & 1 & - & \multicolumn{1}{r|}{-} & 1 & 3 & 1 & - & - & \multicolumn{1}{r|}{1}  & -  & 1  & -  & - & -  & - \\ \hline
\textbf{UR3} & 2 & 3 & \multicolumn{1}{r|}{3} & - & - & \multicolumn{1}{r|}{-} & 2 & 1 & - & - & - & \multicolumn{1}{r|}{-}  & 4  & 3  & -  & 9 & 7  & 5 \\ \hline
\textbf{UR4} & 1 & 3 & \multicolumn{1}{r|}{1} & - & - & \multicolumn{1}{r|}{-} & - & 1 & 1 & 3 & 5 & \multicolumn{1}{r|}{2}  & 36 & 37 & 35 & 1 & -  & - \\ \hline
\textbf{UR5} & 2 & - & \multicolumn{1}{r|}{-} & - & - & \multicolumn{1}{r|}{-} & - & - & - & 4 & 5 & \multicolumn{1}{r|}{-}  & 59 & 59 & 64 & 11 & 7  & 3 \\ \hline
\end{tabular}}

%% file: appendix/annotation_guidelines.tex
\subsection{Annotation Task}
Learning from textual user feedback, i.e., free-form text feedback that expresses user dissatisfaction, describes new knowledge (new concepts), provides corrections or alternative responses, is increasingly important to continuously improve dialog systems. As of today, only few datasets provide annotation for such data. However, there are various dialog datasets publicly available -- from a variety of domains and for all kinds of tasks. As an annotator, you are provided with two different taxonomies, an error type taxonomy, and a user response type taxonomy, and two different types of dialogs, conspicuous dialogs and cold dialogs. Your task is to annotate the dialogs using the given taxonomies. The conspicuous dialogs are identified to contain conspicuous sentences in user utterance that address an error in the previous system response. You have to decide (1) whether or not this conspicuous sentence correctly addresses an error in the previous system response, (2) which error type is represented in the previous system response, and (3) how the user reacted. The cold dialogs do not contain such information. In this case, your task is to identify error situations and user response types in the dialogs just given the taxonomies.

\subsection{Dialog Format}~\label{sec:dialogue_example}

To avoid bias, we provide the dialogs in a unified format without the original task-specific annotations. Here's an example of a conspicuous dialog:

\begin{lstlisting}[linewidth=\columnwidth,breaklines=true,showstringspaces=false,language=Python]
{
    "unique_id": "PMUL0121.json_multiwoz_train",
    "id": "PMUL0121.json",
    "turns": [
        {
            "utterance": "Hi there! Can you give me some info on Cityroomz?",
            "response": "Cityroomz is located at Sleeperz Hotel, Station Road in the Centre area. It is a moderatley priced hotel with a 0 star rating. Would you like to book a room?"
        },
        {
            "utterance": "Yes please. I need it for 7 people for 3 nights starting on tuesday.",
            "response": "How many days would you like to book it for?"
        },
        {
            "utterance": "I would like to book it for 3 nights beginning on Tuesday.",
            "response": "Ok, Your rooms have been booked. Your reference number is CYJDAR7R . Is there anything else I can help you with today?"
        },
        {
            "utterance": "Yeah, I need a restaurant in the west and with expensive pricing.",
            "response": "There are 9 expensive restaurants in the west area. Do you prefer a certain type of food?"
        },
        {
            "utterance": "No, I don't have a preference, but I need the phone number and postcode.",
            "response": "I recommend Graffiti, the number is 01223277977 and the postcode is cb30lx. Would you like me to book a table for you?"
        },
        {
            "utterance": "No, right now, I just need the infor. Thanks!",
            "response": "Can I help you with anything else today?"
        },
        {
            "utterance": "That is all. thanks",
            "response": "Thank you for using our system."
        }
    ],
    "conspicuous": [
        {
            "phrase": [
                "actually, i apologize no need to book, i was just gathering information.", " i would like to book it for 3 nights beginning on tuesday."
            ],
            "confidence": 0.593,
            "turn": 2
        }
    ],
    "annotations": [
        {
            "turn": 2,
            "annotation": {
                "error_type": "E2",
                "comment": "the system misses intent/slots. the user already said that he need it for three nights",
                "error": "C1",
                "user_response": "B3"
            }            
        }
    ]
}

\end{lstlisting}

Each dialog consists of a unique id, an id, and its turns. 
Conspicuous is an array. The first value is an error-indicating phrase, a phrase that was identified to express user dissatisfaction in the utterance of the corresponding turn. The second value is the value from an utterance of this dialog  that was identified to be similar to this error-indicating sentence. Confidence represents the similarity. Dialogs with multiple conspicuous values are possible. The annotations list has an entry for each conspicuous phrase. Please add your annotations here. In comment, you can share your thoughts with us.

Here's an example for an cold dialog:

\begin{lstlisting}[linewidth=\columnwidth,breaklines=true,showstringspaces=false,language=Python]
[
    {
        "dialog": "p2 cats are like cartoons. p1 that's cool , whats your favorite food ? p2 pizza. p1 ni hao . as my father says . you must have great plans ahead ? p2 yes, i plan to be a success.",        
        "error": "C2",
        "error_type": "",
		"user_response": "",
		"comment": "",
        "turn": "",
        "phrase": "",        
    },    
    ...
]
\end{lstlisting}

The structure is a bit different. All cold dialogs are provided in one large json file, and the dialogs themselves maintain the structure of the original dataset. In this case, it is an dialog from the human-bot split of the Self-Feeding Chatbot (p2 represents the system, p1 represents the user). There are two additional fields here: turn and phrase. If you by chance find a phrase that indicates dissatisfaction in the user's response to a system's error, please add phrase and turn to these fields. 

\subsection{Taxonomies}

\subsubsection{Error Type Taxonomy}

This is the taxonomy for the field error type.

\begin{lstlisting}[linewidth=\columnwidth,breaklines=true,showstringspaces=false,language=Python]
[
	{
		"id": "E1",		
		"name": "Ignore Question",
		"description": "The response ignores a user\'s question.",
        "example": "User: 'How is the weather in Berlin?', System:'Have you ever heard of Bobby Jean?'"
	},
	{
		"id": "E2",		
		"name": "Ignore Request",
		"description": "The response ignores a user's request to do something.",
        "example": "User: 'Can you book me a room for two please?', System: 'Look at these flowers... Aren't they beautiful?'"
	},
	{
		"id": "E3",		
		"name": "Ignore Expectation",
		"description": "The response does not fulfill the user's expectation.",
        "example": "User: 'Can you book me a room at The Inn for two nights starting tomorrow? Can you also provide me with the address?', System:'Absolutely!'"
	},
	{
		"id": "E4",		
		"name": "Attribute Error",
		"description": "The response suggests that the system did not get the attributes/slots right.",
        "example": "User: 'Can you book me an expensive restaurant in the west of town?', System:'Sure. I would recommend The Breakfast Club. It is a very cheap and nice location."
	},
	{
		"id": "E5",		
		"name": "Factually Incorrect",
		"description": "The response contains information that is factually incorrect.",
        "example": "User: 'Who is the current president of the United States of America?', System:'Abraham Lincoln.'"
	},
	{
		"id": "E6",		
		"name": "Topic Transition Error",
		"description": "The response transitions to another / a previous topic without reasonable explanation.",
        "example": "User: 'I think I have a cold.', System:'I really love pears. So juicy!'"
	},
	{
		"id": "E7",		
		"name": "Conversationality",
		"description": "The response indicates that the system lost track, e.g., it repeats previous responses (without asking for missing information) or contradicts itself.",
        "example": "System: 'Man is shaped by his environment.', User:'Yes. But does that justify this tone?', System: 'Man is shaped by his environment.'"
	},
	{
		"id": "E8",		
		"name": "Unclear Intention",
		"description": "The response suggests that the user's intent was not successfully conveyed.",
        "example": "User: 'Can you book me a flight to London?', System: 'Yes. What do you want for dinner?'"
	},
	{
		"id": "E9",		
		"name": "Lack of Sociality",
		"description": "The response lacks consideration of social standards, e.g., greetings, is toxic or disrespectful.",
        "examples": "User: 'Hi! My name is Jochen Schweitzer!', System: 'Sure. How can I help you?'"
	},
	{
		"id": "E10",		
		"name": "Lack of Common Sense",
		"description": "The information in the responses opposes the opinion of the majority.",
        "example": "User: 'You have to work hard for success.', System: 'Hmm... No. What is to be, will be.'"
	}	
]
\end{lstlisting}

\subsubsection{User Response Taxonomy}

This is the taxonomy for the field user response.

\begin{lstlisting}[linewidth=\columnwidth,breaklines=true,showstringspaces=false,language=Python]
[
	{
		"id": "UR1",
		"short": "The user ignores the error and continues the conversation.",
		"description": "The user simply continues and does not draw the system's attention to the error.",
		"example": "-"
	},
	{
		"id": "UR2",
		"short": "The user repeats or rephrases his/her concern.",
		"description": "The user repeats or rephrases his originally concern.",
		"example": "'Can you book a restaurant for two for tonight?' vs. 'Can you book a table for two for tonight?'"
	},
	{
		"id": "UR3",
		"short": "The user makes the system aware of the error and provides a correction.",
		"description": "The user makes the system aware of the error and provides information to address what is missing or wrong in its utterance. ",
		"example": "'No, I didn't want you to book a table. I just wanted the address!'"
	},
	{
		"id": "UR4",
		"short": "The user makes the system aware without providing a correction.",
		"description": "The user makes the system aware without providing additional information",
		"example": "'No. You're wrong.'"
	},
	{
		"id": "UR5",
		"short": "The user asks for clarification.",
		"description": "The user is puzzled and asks for clarification, e.g. the system suddenly switches to another topic or mixed concepts up.",
		"example": "'What do you mean?'"
	}
]
\end{lstlisting}

%% file: appendix/experiments.tex
\paragraph{Hyperparameters} All baseline models were trained for five epochs. For the experiment using erroneous dialogs, we trained the models for ten epochs. We used a batch size of 32 and a learning rate of $5e-5$ with no warmup steps. As optimizer, we used the implementation of AdamW~\footnote{\href{https://pytorch.org/docs/stable/generated/torch.optim.AdamW.html}{AdamW} in the Pytorch documentation (last accessed 30. September 2023).}~\citep{loshchilov2019decoupled} in Pytorch. Except for LLAMA~\citep{touvron2023llama}, we fully-finetuned all models. For LLAMA, we only finetuned the LoRA~\citep{hu2021lora} weights, using a rank of 8, an alpha of 16, and a dropout rate of 0.05.

\paragraph{Results} Table~\ref{tab:baseline_experiments} shows the results of our baseline experiments using word-overlapping F1-Score and BLEU~\citep{papineni-etal-2002-bleu}.

\begin{table}[ht]
  \centering
  \input{tables/baseline_experiments}
\caption{Results of our baseline experiments.}
\label{tab:baseline_experiments}
  \end{table}

Compared to the results of our experiment using erroneous dialogs (Table~\ref{tab:continuous_experiments}), both F1-Score and BLEU are higher in case of GPT-2 and LLAMA. We attribute this to the comparatively bad quality of the erroneous dialogs. The majority of these dialogs comes from the SFC~\citep{hancock-etal-2019-learning} dataset, where most errors are due to out-of-context system utterances (see Section~\ref{sec:error_analysis}) and were ignored by users (see Section~\ref{sec:user_reactions}).

%% file: tables/baseline_experiments.tex
\resizebox*{0.45\linewidth}{!}{\begin{tabular}{crr}
\textbf{} & \multicolumn{1}{c}{\textbf{F1}} & \multicolumn{1}{c}{\textbf{BLEU}} \\ \hline
\multicolumn{1}{c|}{Flan-T5} & 56.3 & 35.7 \\ \hline
\multicolumn{1}{c|}{GPT-2} & 47.7 & 36.6 \\ \hline
\multicolumn{1}{c|}{LLAMA} & 57.9 & 42.1
\end{tabular}}

%% file: appendix/polite_answers.tex
We find that it hard to identify errors in human-human dialogs, as the tone is in principle mostly polite, even if one of the partners is wrong. The following dialog is an example from WoW~\citep{dinan2018wizard}:

\begin{itemize}
    \item [\textbf{1. USER:}] I adore photography even though I'm bad at it. It was never a good idea to give me access to a DSLR camera...
    \item [\textbf{1. SYSTEM:}] Well, photography is a science. A little practice and you'd get it down.
    \item [\textbf{2. USER:}] You think so? I think most people, myself included, would consider in an art. But I suppose it's true that there is a lot of technical detail to it.
    \item [\textbf{2. SYSTEM:}] I personally do it as a hobby but many do in businesses, film ect. In my opinion it's always art. i love it because it can capture what your heart see's. If you angle it just right    
\end{itemize}

Here, one partner claims that photography is science. However, it's commonly considered as a visual art. Instead of directly correcting him, the other person rather suggests that the system might be wrong. The following example is from PC~\citep{zhang-etal-2018-personalizing}:

\begin{itemize}
    \item [\textbf{1. USER:}] Hi, how it is going today?
    \item [\textbf{1. SYSTEM:}] It is going great today; everyday is great in my ford truck.
    \item [\textbf{2. USER:}]  Oh. I am eating a piece of fried chicken.
    \item [\textbf{2. SYSTEM:}] Right on. I do not allow fried chicken in my ford truck.
    \item [\textbf{3. USER:}] What do you do for a living?
    \item [\textbf{3. SYSTEM:}] I work on ford trucks. Do you have a ford truck?
    \item [\textbf{4. USER:}] No , i drive a toyota.
    \item [\textbf{4. SYSTEM:}] I am so sorry. You should get yourself a ford truck. Mine is black .
    \item [\textbf{5. USER:}] I am not. I love my toyota .
    \item [\textbf{5. SYSTEM:}] Well okay. How about a ford car? They are just as good.
    \item [\textbf{6. USER:}] Nope, hardcore toyota family. My wife drives a 4runner.
    \item [\textbf{6. SYSTEM:}] Why do they call it a 4runner?
    \item [\textbf{7. USER:}] Because it will run forever.
    \item [\textbf{7. SYSTEM:}] Oh, I thought it was because 4 wheeler was taken.
    \item [\textbf{8. USER:}] Oh. That does not make much sense.	
    \item [\textbf{8. SYSTEM:}] Are you kidding me? Do you know any jokes?
\end{itemize}

In the second to last turn, one partner makes a joke that the other does not understand. The response is quite polite. He informs the partner that his statement does not make much sense to him.

%% file: paper.bbl
\begin{thebibliography}{32}
\expandafter\ifx\csname natexlab\endcsname\relax\def\natexlab#1{#1}\fi

\bibitem[{Bordes et~al.(2017)Bordes, Boureau, and Weston}]{bordes2016learning}
Antoine Bordes, Y-Lan Boureau, and Jason Weston. 2017.
\newblock \href {https://openreview.net/forum?id=S1Bb3D5gg} {Learning
  end-to-end goal-oriented dialog}.
\newblock In \emph{International Conference on Learning Representations}.

\bibitem[{Budzianowski et~al.(2018)Budzianowski, Wen, Tseng, Casanueva, Ultes,
  Ramadan, and Ga{\v{s}}i{\'c}}]{budzianowski-etal-2018-multiwoz}
Pawe{\l} Budzianowski, Tsung-Hsien Wen, Bo-Hsiang Tseng, I{\~n}igo Casanueva,
  Stefan Ultes, Osman Ramadan, and Milica Ga{\v{s}}i{\'c}. 2018.
\newblock \href {https://doi.org/10.18653/v1/D18-1547} {{M}ulti{WOZ} - a
  large-scale multi-domain {W}izard-of-{O}z dataset for task-oriented dialogue
  modelling}.
\newblock In \emph{Proceedings of the 2018 Conference on Empirical Methods in
  Natural Language Processing}, pages 5016--5026, Brussels, Belgium.
  Association for Computational Linguistics.

\bibitem[{Christiano et~al.(2017)Christiano, Leike, Brown, Martic, Legg, and
  Amodei}]{christiano2023deep}
Paul~F Christiano, Jan Leike, Tom Brown, Miljan Martic, Shane Legg, and Dario
  Amodei. 2017.
\newblock \href
  {https://proceedings.neurips.cc/paper_files/paper/2017/file/d5e2c0adad503c91f91df240d0cd4e49-Paper.pdf}
  {Deep {Reinforcement} {Learning} from {Human} {Preferences}}.
\newblock In \emph{Advances in {Neural} {Information} {Processing} {Systems}},
  volume~30. Curran Associates, Inc.

\bibitem[{Chung et~al.(2022)Chung, Hou, Longpre, Zoph, Tay, Fedus, Li, Wang,
  Dehghani, Brahma, Webson, Gu, Dai, Suzgun, Chen, Chowdhery, Narang, Mishra,
  Yu, Zhao, Huang, Dai, Yu, Petrov, Chi, Dean, Devlin, Roberts, Zhou, Le, and
  Wei}]{flan_t5}
Hyung~Won Chung, Le~Hou, Shayne Longpre, Barret Zoph, Yi~Tay, William Fedus,
  Eric Li, Xuezhi Wang, Mostafa Dehghani, Siddhartha Brahma, Albert Webson,
  Shixiang~Shane Gu, Zhuyun Dai, Mirac Suzgun, Xinyun Chen, Aakanksha
  Chowdhery, Sharan Narang, Gaurav Mishra, Adams Yu, Vincent Zhao, Yanping
  Huang, Andrew Dai, Hongkun Yu, Slav Petrov, Ed~H. Chi, Jeff Dean, Jacob
  Devlin, Adam Roberts, Denny Zhou, Quoc~V. Le, and Jason Wei. 2022.
\newblock \href {https://doi.org/10.48550/ARXIV.2210.11416} {Scaling
  instruction-finetuned language models}.

\bibitem[{Dinan et~al.(2019)Dinan, Roller, Shuster, Fan, Auli, and
  Weston}]{dinan2018wizard}
Emily Dinan, Stephen Roller, Kurt Shuster, Angela Fan, Michael Auli, and Jason
  Weston. 2019.
\newblock \href {https://openreview.net/forum?id=r1l73iRqKm} {Wizard of
  wikipedia: Knowledge-powered conversational agents}.
\newblock In \emph{International Conference on Learning Representations}.

\bibitem[{Dybkjaer et~al.(1996)Dybkjaer, Bernsen, and
  Dybkjaer}]{dybkjaer-etal-1996-grice}
Laila Dybkjaer, Niels~Ole Bernsen, and Hans Dybkjaer. 1996.
\newblock \href {https://aclanthology.org/C96-1056} {{G}rice incorporated:
  Cooperativity in spoken dialogue}.
\newblock In \emph{{COLING} 1996 Volume 1: The 16th International Conference on
  Computational Linguistics}.

\bibitem[{Hancock et~al.(2019)Hancock, Bordes, Mazare, and
  Weston}]{hancock-etal-2019-learning}
Braden Hancock, Antoine Bordes, Pierre-Emmanuel Mazare, and Jason Weston. 2019.
\newblock \href {https://doi.org/10.18653/v1/P19-1358} {Learning from dialogue
  after deployment: Feed yourself, chatbot!}
\newblock In \emph{Proceedings of the 57th Annual Meeting of the Association
  for Computational Linguistics}, pages 3667--3684, Florence, Italy.
  Association for Computational Linguistics.

\bibitem[{Henderson et~al.(2019)Henderson, Budzianowski, Casanueva, Coope,
  Gerz, Kumar, Mrk{\v{s}}i{\'c}, Spithourakis, Su, Vuli{\'c}, and
  Wen}]{henderson-etal-2019-repository}
Matthew Henderson, Pawe{\l} Budzianowski, I{\~n}igo Casanueva, Sam Coope,
  Daniela Gerz, Girish Kumar, Nikola Mrk{\v{s}}i{\'c}, Georgios Spithourakis,
  Pei-Hao Su, Ivan Vuli{\'c}, and Tsung-Hsien Wen. 2019.
\newblock \href {https://doi.org/10.18653/v1/W19-4101} {A repository of
  conversational datasets}.
\newblock In \emph{Proceedings of the First Workshop on NLP for Conversational
  AI}, pages 1--10, Florence, Italy. Association for Computational Linguistics.

\bibitem[{Higashinaka et~al.(2021)Higashinaka, Araki, Tsukahara, and
  Mizukami}]{higashinaka-etal-2021-integrated}
Ryuichiro Higashinaka, Masahiro Araki, Hiroshi Tsukahara, and Masahiro
  Mizukami. 2021.
\newblock \href {https://aclanthology.org/2021.sigdial-1.10} {Integrated
  taxonomy of errors in chat-oriented dialogue systems}.
\newblock In \emph{Proceedings of the 22nd Annual Meeting of the Special
  Interest Group on Discourse and Dialogue}, pages 89--98, Singapore and
  Online. Association for Computational Linguistics.

\bibitem[{Hu et~al.(2022)Hu, yelong shen, Wallis, Allen-Zhu, Li, Wang, Wang,
  and Chen}]{hu2021lora}
Edward~J Hu, yelong shen, Phillip Wallis, Zeyuan Allen-Zhu, Yuanzhi Li, Shean
  Wang, Lu~Wang, and Weizhu Chen. 2022.
\newblock \href {https://openreview.net/forum?id=nZeVKeeFYf9} {Lo{RA}: Low-rank
  adaptation of large language models}.
\newblock In \emph{International Conference on Learning Representations}.

\bibitem[{Kim et~al.(2022)Kim, Hessel, Jiang, Lu, Yu, Zhou, Bras, Alikhani,
  Kim, Sap, and Choi}]{kim2022soda}
Hyunwoo Kim, Jack Hessel, Liwei Jiang, Ximing Lu, Youngjae Yu, Pei Zhou,
  Ronan~Le Bras, Malihe Alikhani, Gunhee Kim, Maarten Sap, and Yejin Choi.
  2022.
\newblock \href {http://arxiv.org/abs/2212.10465} {Soda: Million-scale dialogue
  distillation with social commonsense contextualization}.

\bibitem[{Krippendorff(2004)}]{krippendorff2004reliability}
Klaus Krippendorff. 2004.
\newblock \href
  {https://doi.org/https://doi.org/10.1111/j.1468-2958.2004.tb00738.x}
  {Reliability in content analysis}.
\newblock \emph{Human Communication Research}, 30(3):411--433.

\bibitem[{Loshchilov and Hutter(2019)}]{loshchilov2019decoupled}
Ilya Loshchilov and Frank Hutter. 2019.
\newblock \href {https://openreview.net/forum?id=Bkg6RiCqY7} {Decoupled weight
  decay regularization}.
\newblock In \emph{International Conference on Learning Representations}.

\bibitem[{M{\"o}ller et~al.(2007)M{\"o}ller, Engelbrecht, and
  Oulasvirta}]{moller2007analysis}
Sebastian M{\"o}ller, Klaus-Peter Engelbrecht, and Antti Oulasvirta. 2007.
\newblock \href
  {https://research.aalto.fi/en/publications/analysis-of-communication-failures-for-spoken-dialogue-systems}
  {Analysis of communication failures for spoken dialogue systems}.
\newblock In \emph{INTERSPEECH 2007, International Speech Communication
  Association (ISCA)}.

\bibitem[{Ouyang et~al.(2022)Ouyang, Wu, Jiang, Almeida, Wainwright, Mishkin,
  Zhang, Agarwal, Slama, Ray, Schulman, Hilton, Kelton, Miller, Simens, Askell,
  Welinder, Christiano, Leike, and Lowe}]{NEURIPS2022_b1efde53}
Long Ouyang, Jeffrey Wu, Xu~Jiang, Diogo Almeida, Carroll Wainwright, Pamela
  Mishkin, Chong Zhang, Sandhini Agarwal, Katarina Slama, Alex Ray, John
  Schulman, Jacob Hilton, Fraser Kelton, Luke Miller, Maddie Simens, Amanda
  Askell, Peter Welinder, Paul~F Christiano, Jan Leike, and Ryan Lowe. 2022.
\newblock \href
  {https://proceedings.neurips.cc/paper_files/paper/2022/file/b1efde53be364a73914f58805a001731-Paper-Conference.pdf}
  {Training language models to follow instructions with human feedback}.
\newblock In \emph{Advances in Neural Information Processing Systems},
  volume~35, pages 27730--27744. Curran Associates, Inc.

\bibitem[{Papineni et~al.(2002)Papineni, Roukos, Ward, and
  Zhu}]{papineni-etal-2002-bleu}
Kishore Papineni, Salim Roukos, Todd Ward, and Wei-Jing Zhu. 2002.
\newblock \href {https://doi.org/10.3115/1073083.1073135} {Bleu: A method for
  automatic evaluation of machine translation}.
\newblock In \emph{Proceedings of the 40th Annual Meeting on Association for
  Computational Linguistics}, ACL '02, page 311–318, USA. Association for
  Computational Linguistics.

\bibitem[{Park et~al.(2021)Park, Li, Patel, Mudgal, Lee, Kim, Matsoukas, and
  Sarikaya}]{park-etal-2021-scalable}
Sunghyun Park, Han Li, Ameen Patel, Sidharth Mudgal, Sungjin Lee, Young-Bum
  Kim, Spyros Matsoukas, and Ruhi Sarikaya. 2021.
\newblock \href {https://doi.org/10.18653/v1/2021.emnlp-main.489} {A scalable
  framework for learning from implicit user feedback to improve natural
  language understanding in large-scale conversational {AI} systems}.
\newblock In \emph{Proceedings of the 2021 Conference on Empirical Methods in
  Natural Language Processing}, pages 6054--6063, Online and Punta Cana,
  Dominican Republic. Association for Computational Linguistics.

\bibitem[{Paszke et~al.(2019)Paszke, Gross, Massa, Lerer, Bradbury, Chanan,
  Killeen, Lin, Gimelshein, Antiga, Desmaison, Köpf, Yang, DeVito, Raison,
  Tejani, Chilamkurthy, Steiner, Fang, Bai, and Chintala}]{paszke2019pytorch}
Adam Paszke, Sam Gross, Francisco Massa, Adam Lerer, James Bradbury, Gregory
  Chanan, Trevor Killeen, Zeming Lin, Natalia Gimelshein, Luca Antiga, Alban
  Desmaison, Andreas Köpf, Edward Yang, Zach DeVito, Martin Raison, Alykhan
  Tejani, Sasank Chilamkurthy, Benoit Steiner, Lu~Fang, Junjie Bai, and Soumith
  Chintala. 2019.
\newblock \href {https://doi.org/10.48550/ARXIV.1912.01703} {Pytorch: An
  imperative style, high-performance deep learning library}.

\bibitem[{Radford et~al.(2019)Radford, Wu, Child, Luan, Amodei, Sutskever
  et~al.}]{gpt2}
Alec Radford, Jeffrey Wu, Rewon Child, David Luan, Dario Amodei, Ilya
  Sutskever, et~al. 2019.
\newblock \href
  {https://d4mucfpksywv.cloudfront.net/better-language-models/language-models.pdf}
  {Language models are unsupervised multitask learners}.
\newblock \emph{OpenAI blog}, 1(8):9.

\bibitem[{Rastogi et~al.(2020)Rastogi, Zang, Sunkara, Gupta, and
  Khaitan}]{rastogi2020towards}
Abhinav Rastogi, Xiaoxue Zang, Srinivas Sunkara, Raghav Gupta, and Pranav
  Khaitan. 2020.
\newblock \href {https://doi.org/10.1609/aaai.v34i05.6394} {Towards scalable
  multi-domain conversational agents: The schema-guided dialogue dataset}.
\newblock \emph{Proceedings of the AAAI Conference on Artificial Intelligence},
  34(05):8689--8696.

\bibitem[{Reimers and Gurevych(2019)}]{reimers-gurevych-2019-sentence}
Nils Reimers and Iryna Gurevych. 2019.
\newblock \href {https://doi.org/10.18653/v1/D19-1410} {Sentence-{BERT}:
  Sentence embeddings using {S}iamese {BERT}-networks}.
\newblock In \emph{Proceedings of the 2019 Conference on Empirical Methods in
  Natural Language Processing and the 9th International Joint Conference on
  Natural Language Processing (EMNLP-IJCNLP)}, pages 3982--3992, Hong Kong,
  China. Association for Computational Linguistics.

\bibitem[{See and Manning(2021)}]{see-manning-2021-understanding}
Abigail See and Christopher Manning. 2021.
\newblock \href {https://aclanthology.org/2021.sigdial-1.1} {Understanding and
  predicting user dissatisfaction in a neural generative chatbot}.
\newblock In \emph{Proceedings of the 22nd Annual Meeting of the Special
  Interest Group on Discourse and Dialogue}, pages 1--12, Singapore and Online.
  Association for Computational Linguistics.

\bibitem[{Shuster et~al.(2022)Shuster, Xu, Komeili, Ju, Smith, Roller, Ung,
  Chen, Arora, Lane, Behrooz, Ngan, Poff, Goyal, Szlam, Boureau, Kambadur, and
  Weston}]{shuster2022blenderbot}
Kurt Shuster, Jing Xu, Mojtaba Komeili, Da~Ju, Eric~Michael Smith, Stephen
  Roller, Megan Ung, Moya Chen, Kushal Arora, Joshua Lane, Morteza Behrooz,
  William Ngan, Spencer Poff, Naman Goyal, Arthur Szlam, Y-Lan Boureau, Melanie
  Kambadur, and Jason Weston. 2022.
\newblock \href {https://doi.org/10.48550/ARXIV.2208.03188} {Blenderbot 3: a
  deployed conversational agent that continually learns to responsibly engage}.

\bibitem[{Song et~al.(2020)Song, Tan, Qin, Lu, and Liu}]{song2020mpnet}
Kaitao Song, Xu~Tan, Tao Qin, Jianfeng Lu, and Tie-Yan Liu. 2020.
\newblock \href
  {https://proceedings.neurips.cc/paper_files/paper/2020/file/c3a690be93aa602ee2dc0ccab5b7b67e-Paper.pdf}
  {{MPNet}: {Masked} and {Permuted} {Pre}-training for {Language}
  {Understanding}}.
\newblock In \emph{Advances in {Neural} {Information} {Processing} {Systems}},
  volume~33, pages 16857--16867. Curran Associates, Inc.

\bibitem[{Touvron et~al.(2023)Touvron, Lavril, Izacard, Martinet, Lachaux,
  Lacroix, Rozière, Goyal, Hambro, Azhar, Rodriguez, Joulin, Grave, and
  Lample}]{touvron2023llama}
Hugo Touvron, Thibaut Lavril, Gautier Izacard, Xavier Martinet, Marie-Anne
  Lachaux, Timothée Lacroix, Baptiste Rozière, Naman Goyal, Eric Hambro,
  Faisal Azhar, Aurelien Rodriguez, Armand Joulin, Edouard Grave, and Guillaume
  Lample. 2023.
\newblock \href {http://arxiv.org/abs/2302.13971} {Llama: Open and efficient
  foundation language models}.

\bibitem[{Ung et~al.(2022)Ung, Xu, and Boureau}]{ung-etal-2022-SaFeRDialogues}
Megan Ung, Jing Xu, and Y-Lan Boureau. 2022.
\newblock \href {https://doi.org/10.18653/v1/2022.acl-long.447}
  {{S}a{F}e{RD}ialogues: Taking feedback gracefully after conversational safety
  failures}.
\newblock In \emph{Proceedings of the 60th Annual Meeting of the Association
  for Computational Linguistics (Volume 1: Long Papers)}, pages 6462--6481,
  Dublin, Ireland. Association for Computational Linguistics.

\bibitem[{Veron et~al.(2021)Veron, Rosset, Galibert, and
  Bernard}]{veron2021evaluate}
Mathilde Veron, Sophie Rosset, Olivier Galibert, and Guillaume Bernard. 2021.
\newblock \href {https://doi.org/10.48550/ARXIV.2102.13589} {Evaluate
  on-the-job learning dialogue systems and a case study for natural language
  understanding}.

\bibitem[{Wolf et~al.(2020)Wolf, Debut, Sanh, Chaumond, Delangue, Moi, Cistac,
  Rault, Louf, Funtowicz, Davison, Shleifer, von Platen, Ma, Jernite, Plu, Xu,
  Le~Scao, Gugger, Drame, Lhoest, and Rush}]{wolf-etal-2020-transformers}
Thomas Wolf, Lysandre Debut, Victor Sanh, Julien Chaumond, Clement Delangue,
  Anthony Moi, Pierric Cistac, Tim Rault, Remi Louf, Morgan Funtowicz, Joe
  Davison, Sam Shleifer, Patrick von Platen, Clara Ma, Yacine Jernite, Julien
  Plu, Canwen Xu, Teven Le~Scao, Sylvain Gugger, Mariama Drame, Quentin Lhoest,
  and Alexander Rush. 2020.
\newblock \href {https://doi.org/10.18653/v1/2020.emnlp-demos.6} {Transformers:
  State-of-the-art natural language processing}.
\newblock In \emph{Proceedings of the 2020 Conference on Empirical Methods in
  Natural Language Processing: System Demonstrations}, pages 38--45, Online.
  Association for Computational Linguistics.

\bibitem[{Xu et~al.(2023)Xu, Ung, Komeili, Arora, Boureau, and
  Weston}]{xu-etal-2023-learning}
Jing Xu, Megan Ung, Mojtaba Komeili, Kushal Arora, Y-Lan Boureau, and Jason
  Weston. 2023.
\newblock \href {https://doi.org/10.18653/v1/2023.acl-long.758} {Learning new
  skills after deployment: Improving open-domain internet-driven dialogue with
  human feedback}.
\newblock In \emph{Proceedings of the 61st Annual Meeting of the Association
  for Computational Linguistics (Volume 1: Long Papers)}, pages 13557--13572,
  Toronto, Canada. Association for Computational Linguistics.

\bibitem[{Zhang et~al.(2018)Zhang, Dinan, Urbanek, Szlam, Kiela, and
  Weston}]{zhang-etal-2018-personalizing}
Saizheng Zhang, Emily Dinan, Jack Urbanek, Arthur Szlam, Douwe Kiela, and Jason
  Weston. 2018.
\newblock \href {https://doi.org/10.18653/v1/P18-1205} {Personalizing dialogue
  agents: {I} have a dog, do you have pets too?}
\newblock In \emph{Proceedings of the 56th Annual Meeting of the Association
  for Computational Linguistics (Volume 1: Long Papers)}, pages 2204--2213,
  Melbourne, Australia. Association for Computational Linguistics.

\bibitem[{Zhang et~al.(2015)Zhang, Zhao, and LeCun}]{zhang2015character}
Xiang Zhang, Junbo Zhao, and Yann LeCun. 2015.
\newblock \href
  {https://proceedings.neurips.cc/paper/2015/file/250cf8b51c773f3f8dc8b4be867a9a02-Paper.pdf}
  {Character-level convolutional networks for text classification}.
\newblock In \emph{Advances in Neural Information Processing Systems},
  volume~28. Curran Associates, Inc.

\bibitem[{Zheng et~al.(2022)Zheng, Sabour, Wen, and Huang}]{zheng2022augesc}
Chujie Zheng, Sahand Sabour, Jiaxin Wen, and Minlie Huang. 2022.
\newblock \href {http://arxiv.org/abs/2202.13047} {Augesc: Large-scale data
  augmentation for emotional support conversation with pre-trained language
  models}.

\end{thebibliography}
